\documentclass[times,twocolumn,final,authoryear,5p]{elsarticle}

\usepackage{lineno,hyperref}

\usepackage{amsmath,amsfonts}
\usepackage{array}
\usepackage[caption=false]{subfig}
\usepackage{textcomp}
\usepackage{stfloats}
\usepackage{url}
\usepackage{verbatim}
\usepackage{graphicx}
\usepackage{cite}
\usepackage{rotating}

\usepackage{graphicx}
\usepackage{times}
\usepackage{amssymb}
\usepackage{blindtext}
\usepackage{comment}
\usepackage{array}
\usepackage{multicol}
\usepackage{multirow}
\usepackage[dvipsnames]{xcolor}
\usepackage{color, colortbl}
\definecolor{Gray}{gray}{0.93}
\definecolor{LGray}{gray}{0.9}
\usepackage{xcolor, soul}

\newcolumntype{L}[1]{>{\raggedright\let\newline\\\arraybackslash\hspace{0pt}}m{#1}}
\newcolumntype{C}[1]{>{\centering\let\newline\\\arraybackslash\hspace{0pt}}m{#1}}

\newcommand{\E}{\mathbb{E}}
\newcommand{\norm}[1]{\left\lVert#1\right\rVert}

\usepackage{algpseudocode}
\usepackage{algorithm}
\newcommand{\ALOOP}[1]{\ALC@it\algorithmicloop\ #1%
  \begin{ALC@loop}}
\newcommand{\ENDALOOP}{\end{ALC@loop}\ALC@it\algorithmicendloop}

\usepackage{algorithmicx}
\algdef{SE}[DOWHILE]{Do}{doWhile}{\algorithmicdo}[1]{\algorithmicwhile\ #1}%
\makeatother
\usepackage{mathtools}
\usepackage{pifont}
\newcommand*\colourcheck[1]{%
  \expandafter\newcommand\csname #1check\endcsname{\textcolor{#1}{\ding{52}}}%
}

\colourcheck{green}
\newcommand{\myparagraph}[1]{\vspace{2pt}\noindent{\bf #1}}
\newcommand*\colourcross[1]{%
  \expandafter\newcommand\csname #1cross\endcsname{\textcolor{#1}{\ding{55}}}%
}

\newcommand*\colourstar[1]{%
  \expandafter\newcommand\csname #1star\endcsname{\textcolor{#1}{\ding{72}}}
}

\colourcross{blue}
\colourcross{red}
\colourstar{blue}
\journal{Journal of \LaTeX\ Templates}

\biboptions{authoryear}

\begin{document}

\begin{frontmatter}

\title{BDA-SketRet: Bi-Level Domain Adaptation for Zero-Shot SBIR}

\author[1]{Ushasi Chaudhuri\corref{cor1}} 
\cortext[cor1]{Corresponding author}
\ead{ushasi@iitb.ac.in}
\author[2]{Ruchika {Chavan}} 
\author[1]{Biplab {Banerjee}}
\author[3]{Anjan {Dutta}}
\author[4]{Zeynep {Akata}}

\address[1]{Indian Institute of Technology Bombay, India}
\address[2]{School of Informatics, The University of Edinburgh, United Kingdom}
\address[3]{Computer Science, University of Exeter, United Kingdom}
\address[4]{Computer Science, University of Tübingen, Germany}

\begin{abstract}
The efficacy of zero-shot sketch-based image retrieval (ZS-SBIR) models is governed by two challenges. The immense distributions-gap between the sketches and the images requires a proper domain alignment. Moreover, the fine-grained nature of the task and the high intra-class variance of many categories necessitates a class-wise discriminative mapping among the sketch, image, and the semantic spaces. 
Under this premise, we propose BDA-SketRet, a novel ZS-SBIR framework performing a bi-level domain adaptation for aligning the spatial and semantic features of the visual data pairs progressively. In order to highlight the shared features and reduce the effects of any sketch or image-specific artifacts, we propose a novel symmetric loss function based on the notion of information bottleneck for aligning the semantic features while a cross-entropy-based adversarial loss is introduced to align the spatial feature maps.
Finally, our CNN-based model confirms the discriminativeness of the shared latent space through a novel topology-preserving semantic projection network. Experimental results on the extended Sketchy, TU-Berlin, and QuickDraw datasets exhibit sharp improvements over the literature. 

\end{abstract}


\end{frontmatter}


\section{Introduction}

Sketches are the most minimalistic representation of visual data. With the advancement in sensor technology, a quick hand-drawn sketch can be used as a query image to retrieve similar category samples from the visual domain. This is especially useful when there is a lack of available visual query sample at hand, while it is only at the mind of the user at a vague picture~\citep{9070216}. In traditional sketch-based image retrieval (SBIR) \citep{eitz2010sketch}, the training and the test classes are the same. However, this is unrealistic as the model may encounter novel classes during inference in many on-the-fly retrieval applications. The {\em zero-shot learning} (ZSL) \citep{xian2017zero, xian2018feature, romera2015embarrassingly} aims to bridge the gap between the non-overlapping sets of training classes (\textit{seen}) and test classes (\textit{unseen}) using semantic side information for visual recognition. 
To this end, attempts have been made to integrate ZSL with SBIR to get the zero-shot sketch-based image retrieval (ZS-SBIR) problem \citep{yelamarthi2018zero}, even without constraining the search space to contain only the unseen classes at test time, e.g., generalized ZS-SBIR (GZS-SBIR) \citep{dutta2020semantically}. This essentially requires aligning the visual and semantic features (which is used for describing the class) in the embedding space. In this paper, we aim to solve the (G)ZS-SBIR problem by looking into some of the existing problems that are explained in the following subsections and propose improved solutions to this end.

 \begin{figure}[t]
 \centering
 {\includegraphics[width=\linewidth]{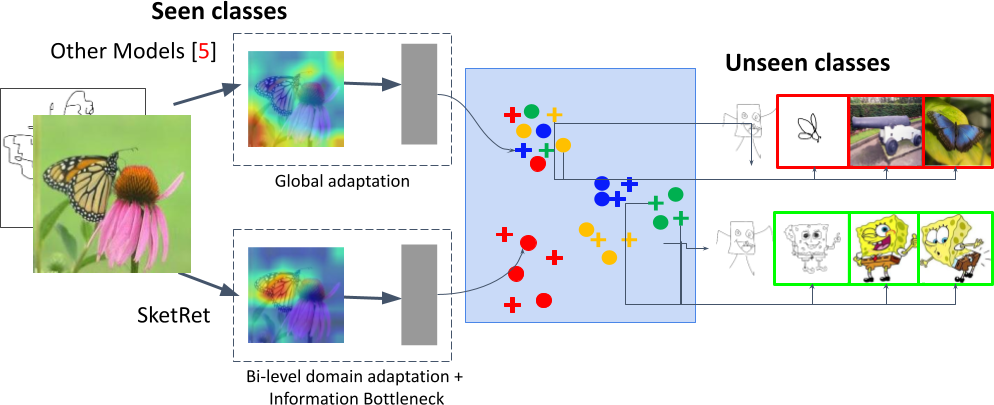}}\caption{ Given a cluttered feature space for images and sketches, a sub-optimal domain alignment fails to produce a discriminative latent space which is affected by problems like hubness and negative transfer. Feature adaptation at multiple feature scales together with a discriminative feature space learning ensures that zero-shot testing can be performed well in BDA-SketRet.} 
 \label{fig:basline}
 \end{figure}

\subsection{Observations}
A key challenge in solving ZS-SBIR is because of the fact that sketches are usually drawn by various artists. ZS-SBIR \citep{yelamarthi2018zero, dutta2020semantically} 
models need to overcome a substantial within-category variance in addition to the domain gap between sketches and images given their disparity in spectral, spatial, and texture properties \citep{xian2017zero, xian2018feature, romera2015embarrassingly}. Moreover, natural images have random background effects which are completely absent in the sketches. In this regard, the majority of the existing ZS-SBIR approaches learn independent mappings to align the visual domains either in a semantic space \citep{dutta2019semantically, p2020stacked} or in a semantics influenced latent space \citep{dey2019doodle}. In both cases, the representations are extracted from the entire image / sketch data while largely neglecting the complex distributions of different local regions. This results in sub-optimal domain alignment as not all the information is equally transferable. Another challenge is that the performance on the unseen classes in  generalized (G)ZS-SBIR is often compromised since the model is biased towards the seen classes. A discriminative visual-semantic mapping is generally preferred to alleviate this issue.  

Further, images and sketches that are significantly dissimilar across domains in the feature space should not be forcefully aligned; otherwise, the model becomes vulnerable to the negative transfer of irrelevant knowledge. This dilemma can be curtailed if the global feature extraction is guided by highly domain-independent mid-level regional properties. The existing work in the ZS-SBIR literature that use domain adaptation primarily tries to align the final features of the sketch and images~\citep{dutta2019semantically, dey2019doodle,duttaadaptive}. This often leads to a sub-optimal domain alignment due to variations in the sketch domain. These methods are non-trivial to use in the ZS-SBIR setup given the disjoint nature of the training and test classes. As the quality of the spatial feature maps obtained from the intermediate layers affects global image-level features extracted from deeper layers in a CNN, it becomes hard to forcefully align the global features without ensuring equivalence of the preceding feature maps. Apparently, the feature maps look substantially different for the sketch and the image modalities considering several issues like the variations in the sketch drawings and background effects in case of images. We argue that the well-aligned mid-level feature maps would supplement the global feature adaptation process. Hence,  we introduce a ZS-SBIR model called BDA-SketRet which inculcates a rigorous multi-level modality adaptation policy between sketches and images both at the mid and high level feature embeddings to ensure a highly domain-agnostic, discriminative, and task-specific representation learning. Fig.~\ref{fig:basline} illustrates in briefs the main idea behind the proposed BDA-SketRet.

\subsection{Our Approach}
 {In BDA-SketRet, we propose to apply cross-modal visual adaptation separately at the local and global levels. The local adaptation is essential as the mid-level feature maps are very dissimilar between the domains and hence making them indistinguishable at this level is challenging. The major problem in aligning the feature maps arises from the vagueness of the decision boundary between the domains. Hence, we propose an cross-entropy based loss measure where the ground truth probability is selected to have high entropy. In the global adaptation, we aim to make the high-level feature embeddings more discriminative and devoid of any irrelevant domain-specific information which may hamper the alignment. For this, we introduce a novel metric learning based information theoretic alignment loss based on symmetric KL divergence. Finally, since the sketch and image modalities are vastly different, we include two cross-modal mid-to-high level feature reconstruction modules which further boosts the domain invariance of the backbone networks. This is different from the traditional cross-reconstruction of the high-level features typically followed in the multi-modal learning tasks.
On the other hand, we propose a semantic projection network for the class prototypes which preserves the neighborhood topology of the original semantic space into the embedding space. Generally, the ZSL models are biased towards to training classes which may partially be solved with neighborhood preservation. In order to accomplish the same, we introduce a network consisting of MLP and graph CNN (GCN). }

 {Our major contributions are as follows: i) We introduce BDA-SketRet, a discriminative multi-level domain adaptation framework for solving ZS-SBIR and GZS-SBIR. ii) We showcase the effectiveness of aligning both the domains in two different feature levels (mid and high level) and propose a novel strategy for the same. A novel symmetric KL divergence based formulation is introduced for information bottleneck loss to be used for semantic feature alignment as opposed to the traditional asymmetric KL based VIB loss measure~\citep{Tishby2015IB}. iii) An intuitive neighborhood-preserving semantic projection module is proposed which is seen to reduce the model bias towards the seen classes specifically for GZS-SBIR.} iv) We conduct extensive experiments on the extended Sketchy, TU-Berlin, and QuickDraw datasets and rigorously ablate the model. The efficacy of the proposed model is verified experimentally as the proposed network beats the existing state-of-the-art models by a margin of $\approx$ 3 - 4\% in all the evaluation metrices. 

\section{Related Works}
{In this section, we review prior related works on ZS-SBIR, semantic projection and GCN that are related to ours. We also show how is our work different from the existing state-of-the-art models.}

\subsection{State-of-the-Art ZS-SBIR Algorithms}
Prior to solving ZS-SBIR tasks, the simple SBIR problem received a lot of attention~\citep{lei2019semi}. As already stated, the major obstacle in solving the SBIR task stems from the fact that the distributions difference between sketch and image data is exceedingly large. Early works in this area include conventional pattern recognition methods for retrieval by engineering hand-crafted visual features~\citep{hu2013performance,saavedra2014sketch}. The proposition behind such approaches is to solve the task at hand by obtaining the edge-map of the natural images and to further match them with sketches arising from the same categories. As expected, the low-level SIFT~\citep{lowe1999object}, SURF~\citep{bay2006surf}, or HoG~\citep{dalal2005histograms} based descriptors are unable to properly encode the regional variations of the sketch data, resulting in an inferior cross-modal matching. The performance measures of deep CNN based SBIR models have witnessed a massive enhancement lately, thanks to the data-driven feature learning capabilities of CNN. Since the retrieval performance benefits from a discriminative feature space,  several endeavors rely on distance metric learning strategies like contrastive-loss~\citep{chopra2005learning}, triplet-loss~\citep{sangkloy2016sketchy}, and HOLEF-based loss~\citep{song2017deep}, to name a few. \citep{bui2017compact} proposed an efficient representation for sketch and image features using triplet loss, while on the other hand \citep{federici2020learning} provides an information-theoretic approach.  Likewise, \citep{song2017deep,qi2016sketch,yu2017sketch,sangkloy2016sketchy,wang2015community,liu2017deep,zhang2018generative} aims to solve a similar task.

{The ZS-SBIR literature consists of both the discriminative and generative deep learning based techniques. Under the generative umbrella, \citep{shen2018zero} proposes a hashing network for the semantic knowledge reconstruction (ZSIH). Similarly, \citep{yelamarthi2018zero} introduces a conditional generative model for ZS-SBIR based on variational learning. The stacked auto-encoder (SAN) method \citep{p2020stacked} deploys a generative framework based on stacked-adversarial networks within a Siamese architecture. The paired cyclic consistency loss proposed in SEM-PCYC~\citep{dutta2019semantically,dutta2020semantically} helps in aligning the sketches and images in an encoded semantic space using adversarial training. Inspired by style transfer, \citep{dutta2019style,dutta2020styleguide} develops a style-guided image to image translation model for ZS-SBIR, while \citep{dey2019doodle} uses a triplet-based network to solve the task at hand. \citep{duttaadaptive} highlights the implications of data and class imbalance in ZS-SBIR and introduces an adaptive margin diversity regularizer (AMD-reg) to combat the same. As opposed to the real-valued feature embedding, hash-code based representations are also considered in this regard which offers a trade-off between performance and storage~\citep{liu2017deep}. The generative models for cross-modal style transfer are also explored~\citep{dutta2019style} in this regard. While all the techniques showcase their performance on ZS-SBIR, a few works~\citep{dutta2019semantically,dutta2020semantically,p2020stacked} also demonstrate their experiments for the GZS-SBIR setting.}  

\subsection{Semantic projection and GCN.}
The semantic information has an important role to play for attaining an improved ZS-SBIR inference. Generally, the visual modalities are aligned to a fixed semantic space. While \citep{dutta2019semantically} uses an encoded semantic space obtained via a semantic auto-encoder, methods of \citep{dey2019doodle, dutta2019style} reconstruct the original semantic vectors from the visual information. Generating visual information from semantic prototypes has been dealt with in this regard \citep{yelamarthi2018zero}. 
The graph CNN \citep{kipf2016semi} can encode the structural similarity among several objects and have been successfully applied for vision tasks like ZSL, image captioning, and visual question answering \citep{kampffmeyer2019rethinking, yang2019auto, narasimhan2018out}.
The graph CNN has been considered in ZS-SBIR recently \citep{zhang2020zero} where the modality alignment takes place by restricting the domain-specific graphs to be isomorphic.

\noindent\textbf{How are we different?} Contrary to the existing literature, we choose to look into the ZS-SBIR problem mainly as a domain adaptation task wherein we propose to do a multi-level adaptation by tackling both spatial and semantic domain drifts. The notion of multi-level domain adaptation is new in this regard, to the best of our knowledge. From the theoretical point of view, we define a new symmetric loss function for the information bottleneck principle which is found to be well-suited for multi-modal labeled data pairs, something that the existing asymmetric divergence based functions fail to model. Finally, our semantic projection network offers topology preservation within minimal overhead and combats hubness effectively where all the existing techniques overlook this aspect completely.


\begin{figure*}[t]
\centering
\includegraphics[width =0.9\linewidth]{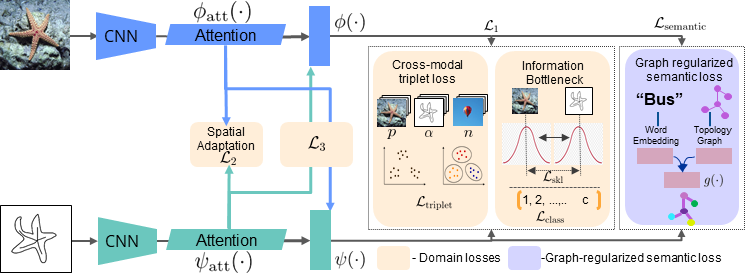}
\caption{A depiction of our BDA-SketRet architecture.The images and sketches undergo a bi-level domain adaptation at the outputs of $(\phi_\text{att}, \psi_\text{att})$ and $(\phi, \psi)$, respectively. The important learning objectives: (i) domain losses, (ii)  cross-modal multi-level reconstruction loss, and (iii) neighborhood  preserving  semantic  projection  loss are shown and how they are evaluated in the network. In the semantic graph, thicker edges means more discriminativeness. In the semantic graph, thicker edges means more discriminativeness. 0/1 represent the domain labels.  
}
\label{fig:block}
\end{figure*}%


\section{Problem definition and preliminaries}\label{sec:sketret}
Let $\mathcal{Z}^s = \{\mathcal{A}^s, \mathcal{B}^s, \mathcal{C}^s, \mathcal{W}^s\}$ be a multi-modal training dataset consisting of images $\mathcal{A}^s$ and sketches $\mathcal{B}^s$ obtained from the $|\mathcal{C}^s|$ \textit{seen} visual categories. Additionally, we have access to semantic side information $\mathcal{W}^s$ which typically corresponds to the distributed word-vector embeddings of the individual category names. During inference, image and sketch data $\mathcal{Z}^u = \{\mathcal{A}^u, \mathcal{B}^u\}$ from a non-overlapping set of previously \textit{unseen} classes $\mathcal{C}^u$ are considered ($\mathcal{C}^{u} \cap \mathcal{C}^{s} = \varnothing $) in the zero shot SBIR setup.
We deal with the unpaired dataset setting in $\mathcal{Z}^s$ where the number of sketch and image instances in $\mathcal{A}^s$ and $\mathcal{B}^s$ are different: $\{a_i^s\}_{i=1}^N \in \mathcal{A}^s$ and $\{b_i^s\}_{i=1}^M \in \mathcal{B}^s$. The model is trained to reduce the distribution mismatch between $\mathcal{A}^s$ and $\mathcal{B}^s$ and subsequently to transfer the knowledge from $\mathcal{Z}^s$ to $\mathcal{Z}^u$ with the help of the semantic information $\mathcal{W}^s$. The testing phase concerns the retrieval of images with similar semantic categories from $\mathcal{A}^u$ given the sketch queries from $\mathcal{B}^u$. In contrast to ZS-SBIR, GZS-SBIR assumes the presence of images from $\mathcal{A}^s \cup \mathcal{A}^u$ during testing for unseen-class sketch queries coming from $\mathcal{B}^u$.


\section{BDA-SketRet}
The goal of BDA-SketRet is to align the images and sketches from the same class in a semantically meaningful shared latent space.
It is composed of 
cross-modal triplets where the sketch data from $\mathcal{B}^s$ serves as the anchor $(\alpha)$ while the positive ($p$) and negative ($n$) counterparts are selected from $\mathcal{A}^s$  (Fig.~\ref{fig:block}).  The feature networks for $\mathcal{A}^s$ and $\mathcal{B}^s$ are defined by $\phi(\cdot)$ and $\psi(\cdot)$ which are convolutional neural networks with integrated attention sub-networks $\phi_\text{att}(\cdot)$ and $\psi_\text{att}(\cdot)$.
The attention block outputs are simultaneously projected to the local adversarial domain classifier $l(\cdot)$ to highlight spatially indistinct features of the same-class samples from $\mathcal{A}^s$ and $\mathcal{B}^s$ and to the shared latent space. We further introduce two cross-modal feature reconstruction modules $(\mathcal{V}_{\alpha}(\cdot),\mathcal{V}_p(\cdot))$ which aim to reconstruct $\phi(p)$ from $\psi_\text{att}(\alpha)$ and $\psi(\alpha)$ from $\phi_\text{att}(p)$ through variational bottlenecks. On the other hand, the outputs of $\phi(\cdot)$ and $\psi(\cdot)$ need to be synchronized for defining the shared embedding space. In this regard, the global domain adaptation on $\phi(p/n)$ and $\psi(\alpha)$ is carried out considering a combination of the domain classifier $f(\cdot)$ and a multi-class category classifier $h(\cdot)$. A semantic sub-network $g(\cdot, \cdot)$ comprising of an MLP $g_1(\cdot)$ and a graph CNN $g_2(\cdot, \cdot)$ is used to non-linearly project the semantic vectors into the shared space.  {The outputs of $g_1(\cdot)$ and $g_2(\cdot,\cdot)$ are concatenated and projected onto the latent space by another MLP $g_3(\cdot)$.}  


There are mainly three learning objectives that together train BDA-SketRet, namely, (i) spatial and semantic domain losses (ii) cross-modal multi-level reconstruction loss, and (iii) neighborhood preserving semantic projection loss. They are detailed in the following subsections.

\subsection{Multi-level Sketch Image Alignment}
The semantic adaptation works on the final features obtained from the last dense layer ($\phi$ and $\psi$) of the backbone, while the spatial adaptation is carried out on the mid-level feature maps obtained from the final convolutional layer ($\phi_{att}$ and $\psi_{att}$). 

\myparagraph{{(i) Semantic Adaptation.}}  We aim to maximise the information between the probability distributions $P_{\alpha} = \text{P}(\psi(\alpha)|\alpha)$ and $ P_{p} = \text{P}(\phi(p)|p)$, and minimise the information they share with $P_{n} = \text{P}(\phi(n)|n)$. We employ a posterior distribution matching method to generate a latent space in which $\alpha$ and $p$ are as aligned as possible, while $\alpha$ and $n$ are sufficiently disjoint.  We model the probability distributions $P_{\alpha}$, $P_{p}$, and $P_{n}$ by appropriate normal distributions.

We follow the deep variational  {information bottleneck framework \citep{45903,federici2020learning}} where the classification task is considered for data from both the visual modalities with the combined cross-entropy loss as $\mathcal{L}_{class}$. The aim is to maximise information between latent variables and output space by minimising cross entropy loss between the distributions, and minimises information between source and latent variables by using the standard normal distribution as a variational approximation to their marginal. We restrain this latent space to be a standard normal distribution.
We minimize the symmetric KL divergence between $P_{\alpha}$ and $P_{p}$ is to reduce their relative entropy. 
 {Apart from reducing the distance between $P_{\alpha}$ and $P_{p}$, our framework also requires that $P_{\alpha}$ and $P_{n}$ be disjoint in the feature space. For this, we define $\mathcal{L}_{\text{skl}}(P_{\alpha}, P_{p/n})$ as $\frac{1}{2} \Big(D_\text{KL}(P_{\alpha}\|P_{p/n}) + D_\text{KL}(P_{p/n}\| P_{\alpha}) \Big)$.} 
Hence, we introduce a novel symmetric KL divergence triplet loss $\mathcal{L}_{\text{t-skl}}$ as, 
\begin{equation}
    \mathcal{L}_{\text{t-skl}} = \beta \max \{0, \mathcal{L}_{\text{skl}}(P_{\alpha}, P_p) -
    \mathcal{L}_{\text{skl}}(P_{\alpha}, P_n) + \lambda \}\label{eq:trip_sym_kl} 
\end{equation}

 Here $\beta$ and $\lambda$ are hyper-parameters which denote the weight of SKL divergence and the margin respectively. By definition of the loss function, our goal is to bring $P_{\alpha}$ and $P_{p}$ closer while maintaining a difference of at least $\lambda$ from the distance between $P_{\alpha}$ and $P_{n}$. 
%

Along with $\mathcal{L}_\text{t-skl}$, we also use instance-based triplet loss $\mathcal{L}_\text{triplet}$ here so that the cross-modal data become class-wise dense clusters in the embedding space.  {They complement each other in obtaining a dense feature space highlighting only the domain-shared features while neglecting any artifact. While $\mathcal{L}_\text{t-skl}$ maximizes the information between latent variables and output space, $\mathcal{L}_\text{triplet}$ makes the cross-modalities dense in the embedding space.} The cross-modal triplet loss aims to bring the same class sample $\phi(p)$ from the image modality closer to a given sketch anchor $\psi(\alpha)$ while pushing the negative image sample $\phi(n)$ far from $\psi(\alpha)$ at least by a margin of $\mu$ as per Eq. \ref{eq:triplet} using the Euclidean distance metric $D$. 
%
%
\begin{equation}
\label{eq:triplet}
\centering
\begin{aligned}
\mathcal{L}_\text{triplet} = \min_{\phi, \psi}  \E_{\alpha \in \mathcal{B}^s, p,n \in \mathcal{A}^s} [\max\{0, \mu +  D(\psi(\alpha), \phi(p)) \\- D(\psi(\alpha), \phi(n))\}] 
\end{aligned}
\end{equation}

\noindent Therefore, the combined semantic adaptation loss is: $ \mathcal{L}_1 = \min_{\phi, \psi, h} \mathcal{L}_\text{t-skl} + \mathcal{L}_\text{triplet}  + \mathcal{L}_\text{class}$.
%

\myparagraph{{(ii) Spatial Adaptation}} encourages the learning of abstract local spatial concepts common to $\mathcal{A}^s$ and $\mathcal{B}^s$. This is done by adversarially adapting the spatially average-pooled feature-map outputs of $\phi_\text{att}(p)$ and $\psi_\text{att}(\alpha)$. 
Most adversarial domain adaptation frameworks employ a binary domain classifier $l(\cdot)$. 
 {We observed that sketches and images are extremely disparate in terms of spatial and spectral properties. Therefore, it is fairly unchallenging for a discriminator to learn distributions corresponding to domains promptly, defeating the purpose of adversarial training. The root of this problem is the hard decision boundaries we set for the domain classifier in the min-max optimisation.} 
We argue that the margin between the decision boundaries is still substantially large for the generator to overcome, given the inherently huge domain gap. Therefore, we introduce a pseudo-decision boundary between spatial features of $\alpha$ and $p$ that helps the generator to surpass the margin. 

 {This decision boundary introduces a certain degree of domain invariance before projecting them onto the latent space, leading to stable training, i.e. $ l(x) = 0.5, \ \ \text{if} \ \ x \in \mathcal{A}^{s}$ or $x \in \mathcal{B}^{s}$. 
In an adversarial setup, the generator competes with the discriminator to increase P$(x\in \mathcal{A}^{s})$ to be greater than 0.5 and P$(x\in \mathcal{B}^{s})$ to be less than 0.5. The decision boundary is symmetrical to both the domains in the feature space, rendering $0.5$ as a good value for a pseudo decision boundary. So the local domain adaptation is given as:}
\begin{equation}
\label{eq:local_minmax}
\centering
\begin{aligned}
\mathcal{L}_2 = \min_{{\phi_\text{att}}, {\psi_\text{att}}} \max_{l} \E_{\alpha \in \mathcal{B}^s, p \in \mathcal{A}^s} [ 0.5 \log (1 - l(\psi_\text{att}(\alpha)))  \\+ 0.5 \log l(\phi_\text{att}(p))] 
\end{aligned}
\end{equation}
%

\myparagraph{(iii) Cross-modal multi-level reconstruction.} Although $\mathcal{L}_2$  adapts the intermediate feature maps of $\mathcal{A}^s$ and $\mathcal{B}^s$, it does not guarantee that the feature maps of a given modality are aware of the final latent feature distributions for the other modality. In this regard, it is essential that the high-level feature distributions $P_{\alpha}$ and $P_{p}$ characterised by domain-independence do not completely lose out class-related information about their conjugate modality i.e. $\alpha$ and $p$. To better equip the feature maps with the cross-modal information, we introduce the classwise cross-modal reconstruction loss using generative modelling. Specifically, the cross-modal encoder-decoder modules $\mathcal{V}_{\alpha}=(\mathcal{V}_{\alpha}^e,\mathcal{V}_{\alpha}^d)$ and $\mathcal{V}_p =  (\mathcal{V}_p^e,\mathcal{V}_p^d)$ reconstruct the latent feature embedding of sketch anchor $\psi(\alpha)$ given the outcome of $\phi_\text{att}(p)$ and vice-versa. 
Both $\mathcal{V}_{\alpha}^e$ and $\mathcal{V}_p^e$ are designed to be stochastic encoders and their outputs follow the standard normal distributions as per the principles of variational learning. Let $D_{\text{KL}}$ be the Kullback-Leibler divergence, referred to as equation~\ref{eq:kullback}. 
\begin{equation}\label{eq:kullback}
    \mathcal{L}_\text{KL}(\mathcal{V},F, x) =   D_\text{KL}(q(\mathcal{V}(F(x)))  \|  \mathcal{N}(0, 1))
\end{equation}
The cross-modal reconstruction loss is given as follows. 
\begin{gather}
     \mathcal{L}_\text{rec}^{1}= \norm{\mathcal{V}_{p}(\phi_\text{att}(p)) - \psi(\alpha)}^{2} + \mathcal{L}_\text{KL}(\mathcal{V}_{p}^{e} , \phi_\text{att}, p)
      \nonumber\\
     \mathcal{L}_\text{rec}^{2}= \norm{\mathcal{V}_{\alpha}(\psi_\text{att}(\alpha)) - \phi(p)}^{2} + 
   \mathcal{L}_\text{KL}(\mathcal{V}_{\alpha}^e,\psi_\text{att},\alpha) \nonumber\\
     \mathcal{L}_3 = \min_{\mathcal{V}_{p}, \mathcal{V}_{\alpha},\phi,\psi} 
     \E_{\alpha \in \mathcal{B}^s, p \in \mathcal{A}^s}   [\mathcal{L}_\text{rec}^{1} +  \mathcal{L}_\text{rec}^{2}]
\end{gather}
The overall domain loss is given by equation~\ref{eq:domainloss}. 
\begin{equation}\label{eq:domainloss}
     \mathcal{L}_\text{domain} = \mathcal{L}_1  + \mathcal{L}_2  + \mathcal{L}_3 
 \end{equation}

%


\subsection{Neighborhood Preserving Semantic Projection Loss}
The semantic side information is obtained by taking various combinations of text-based and hierarchical word embeddings for the category names.
For the distributed word-vector models, we consider the pre-trained  Word2Vec~\citep{mikolov2013efficient} and fasttext~\citep{bojanowski2017enriching} while the Jiang-Conrath~\citep{jiang1997semantic} and path similarity are used for the latter.

As aforementioned, the neighborhood information is severely affected when the semantic prototypes are directly projected onto the latent space using non-linear dense layers, limiting the overall performance. Hence, we desire to ensure the latent space to mimic the neighborhood information of the original semantic space, in addition to being discriminative.
We project the semantic prototypes $\mathcal{W}^s$ together with the topology information of the original semantic space to the shared latent space. The topology information, which is found to bring in a regularization effect into the latent space is encapsulated in the weighted semantic adjacency matrix $\Gamma_{|\mathcal{C}^s| \times |\mathcal{C}^s|}$ defined by the pairwise cosine dissimilarity among the semantic prototypes of the seen classes. Ideally, the outputs of $g_1(\mathcal{W}^s)$ and $g_2(\Gamma, \mathcal{W}^s)$ are concatenated and subsequently projected onto the latent space by another MLP $g_3(\cdot)$: $g(\mathcal{W}^s, \Gamma) = g_3([g_1(\mathcal{W}^s), g_2(\Gamma, \mathcal{W}^s)])$ where $[\cdot,\cdot]$ defines the vector concatenation operation. 
The semantic reconstruction loss brings $\phi(p)$ and $\psi(\alpha)$ closer to the projected class embedding $g(w^+, \Gamma)$ while maximizing the divergence between $\phi(n)$ and  $g(w^+, \Gamma)$. This is accomplished through the graph regularized semantic loss $\mathcal{L}_\text{semantic}$ as follows,
\begin{equation*}
\centering
\begin{aligned}
\mathcal{L}_\text{semantic} = \min_{g,\phi,\psi} \E_{\alpha \in \mathcal{B}^s, p,n \in \mathcal{A}^s,w^+ \in \mathcal{W}^s} [S(\psi(\alpha), g(w^{+}, \Gamma), 1) \\ +  S(\phi(p), g(w^{+}, \Gamma), 1)  +  S(\phi(n), g(w^{+}, \Gamma), 0)]
\end{aligned}
\end{equation*}

\noindent where the distance $S$ between the vectors $(\textbf{x}, \textbf{y})$ is defined in terms of the cosine distance for a given threshold $t$ as, $S(\textbf{x}, \textbf{y}, t) = \frac{1}{2} ( t - \frac{\textbf{x} \textbf{y}^{T}}{\norm{\textbf{x}} \norm{\textbf{y}}} )$.
%
%
The overall objective function for the BDA-SketRet framework can now be put forward as $\mathcal{L}_\text{total} = \mathcal{L}_\text{domain} +  \mathcal{L}_\text{semantic} $.
%

\section{Theoretical proof for IB-based alignment for tighter generalization bound}
In this section, we provide a theoretical proof to show that the proposed semantic adaptation and information bottleneck framework generate a latent space with two properties: the projections of $\alpha$ and $p$ are coalesced, while those of $\alpha$ and $n$ are disjoint. To prove this, we show that the distance between probability distributions of $\alpha$ and $n$ is always greater than the distance between $\alpha$ and $p$ using the theory of learning from different domains.

Adopting the notation from \citep{BenDavid2009ATO}, we denote error given by a hypothesis $h \in \mathcal{H}$  on a source domain $\mathcal{S}$ and target domain $\mathcal{T}$ by $\epsilon_{\mathcal{S}}$ and $\epsilon_{\mathcal{T}}$ respectively, where $\mathcal{H}$ defines the hypothesis space. The error is defined as the probability according to the distribution $D$ that a hypothesis $h$ disagrees with a true labeling function $f$. 

Let us first consider the objective of generating fused $P_{\alpha}$ and $P_{p}$. Without any loss of generality, we fix the source domain to be $D_{\alpha} = \{ x\in A^{s} \}$
and target domain to $D_{p} = \{ x\in B^{s}  |$  $x$ $\text{belongs to class}$ $w^{+} \}$. Translating our notations to the theoretical bound obtained in \citep{BenDavid2009ATO}, we provide the bound for the error on target domain as
\begin{equation}
\label{eq:bound}
    \centering
    \epsilon_p \leq d_{\mathcal{H}\Delta \mathcal{H}}(P_{\alpha}, P_{p}) + \epsilon_\alpha(h) + \underset{h' \in \mathcal{H}}{\text{min}} \epsilon_p(h') + \epsilon_\alpha(h')  
\end{equation}

The constraints on the latent distributions to be the standard normal distribution as per the variational upper bound principle, results in the following inequality by using the triangular inequality for the $d$-divergence.
\begin{equation}
\label{eq:tri_1}
    d_{\mathcal{H}} (P_{\alpha},P_{p}) \leq d_{\mathcal{H}} (P_{\alpha}, \mathbb{N}(0,\mathbb{I})) + d_{\mathcal{H}} (P_{p}, \mathbb{N}(0,\mathbb{I}))
\end{equation}

Similarly, by virtue of the information bottleneck, we can again use the triangle inequality for $ d_{\mathcal{H}} (P_{\alpha},P_{p})$ as
\begin{equation}
\label{eq:tri_2}
    d_{\mathcal{H}} (P_{\alpha},P_{n}) \leq d_{\mathcal{H}} (P_{\alpha}, \mathbb{N}(0,\mathbb{I})) + d_{\mathcal{H}} (P_{n}, \mathbb{N}(0,\mathbb{I}))
\end{equation}

By Subtracting Eq. \ref{eq:tri_1} from \ref{eq:tri_2}, we get
\begin{gather}
\label{opt}
    d_{\mathcal{H}} (P_{\alpha},P_{n}) - d_{\mathcal{H}} (P_{\alpha},P_{n}) \leq d_{\mathcal{H}} (P_{p}, \mathbb{N}(0,\mathbb{I})) - \nonumber \\  d_{\mathcal{H}} (P_{n}, \mathbb{N}(0,\mathbb{I})) 
\end{gather}

We would like to point out that the RHS consists of terms that are minimized by the variational bottleneck framework. Therefore from Eq. \ref{opt} , under optimal conditions on $\mathcal{L}_{\text{t-skl}}$, $d_{\mathcal{H}} (P_{\alpha},P_{p}) - d_{\mathcal{H}} (P_{\alpha},P_{n}) \leq 0$. Therefore, we can conclude that the generated latent space probability distributions follow $d_{\mathcal{H}} (P_{\alpha},P_{n}) \geq d_{\mathcal{H}} (P_{\alpha},P_{p})$. 

{
\setlength{\tabcolsep}{4pt}
\renewcommand{\arraystretch}{1}
\begin{table*}\caption{{Comparing our BDA-SketRet with SOTA on ZS-SBIR (top) and GZS-SBIR (bottom) on both the splits of Sketchy-extended (S1 and S2), TU Berlin-extended and Quickdraw-extended datasets. All models use VGG-16 feature backbone. The '-' represents the evaluation metrices which were not mentioned in the respective papers.}}
\label{tab:comparison}
\centering\resizebox{\linewidth}{!}{
\begin{tabular}{L{6.3cm}|cc|cc|cc|cc}
     &
    \multicolumn{2}{c}{\textbf{Sketchy-ext}  (S2)}&
     \multicolumn{2}{c}{\textbf{Sketchy-ext}  (S1)}&
    \multicolumn{2}{c}{\textbf{TU Berlin-ext} } & 
    \multicolumn{2}{c}{\textbf{Quickdraw-ext} }\\
     \textbf{Model} & \textbf{P@ 200}& \textbf{mAP @200}&
     \textbf{mAP} &\textbf{P@ 100}&
     \textbf{mAP} &\textbf{P@ 100}&
     \textbf{mAP} &\textbf{P@ 100}\\
    \hline  
ZSIH~\citep{shen2018zero} &-& -& -&-&22.0& 29.1& 13.1& 18.8\\
CVAE~\citep{yelamarthi2018zero}&19.6 &-& 33.3& 22.5& 00.5& -&- &-\\  
ZS-SBIR~\citep{yelamarthi2018zero} &- &- &19.6&28.4&00.5&00.1 &00.6&00.1\\
SEM-PCYC~\citep{dutta2019semantically} &37.0 &45.9 &34.9&46.3&29.7&42.6&\textbf{17.7}&25.5\\ 
Doodle2search~\citep{dey2019doodle}&  37.0& 46.1 &-&-&10.9 &- &  07.5&-\\ 
Style-guide \citep{dutta2019style}&40.0&35.8&37.6&48.4&25.4 &35.5 &-&-\\
Style-guide+AMDReg \citep{duttaadaptive}&-&-&41.0&51.2&29.1 &37.6 &-&-\\
SEM-PCYC+AMDReg \citep{duttaadaptive} &-&-&39.7&49.4&33.0&47.3 &-&-\\
\textbf{BDA-SketRet (Ours)} & \textbf{45.8}& \textbf{55.6}&\textbf{43.7}&\textbf{51.4}&\textbf{37.4}&\textbf{50.4}&15.4&\textbf{44.0}\\
\hline
ZS-SBIR~\citep{yelamarthi2018zero}&-&-&14.6&19.0&00.3&00.1&00.2&00.1\\
SEM-PCYC \citep{dutta2019semantically}&-&-&30.7&36.4&19.2&29.8&14.0&22.1\\
SEM-PCYC+AMDReg \citep{duttaadaptive}&-&-&32.0&39.8&{24.5}&30.3&-&-\\
Style-guide \citep{dutta2019style}&-&-&33.0&38.1&14.9& 22.6 &-&- \\
\textbf{GZS-BDA-SketRet (Ours) }&\textbf{22.6} &  \textbf{33.7}&\textbf{33.8}&\textbf{41.3}&\textbf{25.1} &\textbf{35.7}& \textbf{15.4}&\textbf{28.6}
\end{tabular}
}
\end{table*}
}

\section{Experiments}
\myparagraph{Datasets.} We validate the efficacy of the BDA-SketRet by performing experiments on the benchmark Sketchy-extended~\citep{sangkloy2016sketchy}, TU Berlin-extended (TUB)~\citep{eitz2012humans}, and the newly introduced QuickDraw-extended~\citep{dey2019doodle} datasets. Sketchy consists of $125$ categories of unpaired sketch and photo images. We use the two conventional train-test splits. In split 1 (S1) we randomly select 25 classes as the unseen test data, while in split 2 (S2) we use $|\mathcal{C}^s| : |\mathcal{C}^u| = 104:21$ as mentioned in~\citep{yelamarthi2018zero} where the $21$ unseen classes are carefully chosen not to be part of ImageNet~\citep{deng2009imagenet}. For the remaining datasets, we follow the same protocol as~\citep{dutta2020semantically}.


\myparagraph{Training and Evaluation Protocol.} 
We select around $5000$ triplets in each training iteration based on the aforementioned triplet-mining protocol. $\mu$ (Eq. \ref{eq:triplet}) and $\lambda$ (Eq.~\ref{eq:trip_sym_kl}) for Sketchy: (0.1, 0.1), Tu-Berlin: (1, 1), and QuickDraw: (1, 1), while $\beta$ is set to 0.0001 for all the datasets. These parameters were estimated using grid search with cross-validation~\citep{yelamarthi2018zero}.  We use batch-normalization and leaky-ReLU non-linearity after each of the layers to ensure a stable training. $\mathcal{L}_\text{total}$ is optimized using the stochastic gradient descent (SGD) with momentum as the optimizer with a mini-batch size of $32$. An initial learning rate of $0.0001$ and a momentum of $0.9$ are set.  We find that $\mathcal{L}_\text{total}$ converges for all the datasets within $50$ epochs. We report the performance of BDA-SketRet in terms of mAP@all (mean average precision), mAP@200, P@100, and P@200, respectively, where P stands for precision.

\myparagraph{Implementation Details.}
The feature backbone networks $\phi(\cdot)$ and $\psi(\cdot)$ are the ImageNet pre-trained VGG-16 model \citep{simonyan2014very}.
Two modality specific spatial attention learning modules consisting of convolution kernels with sigmoid non-linearity are applied on the outputs of the final convolution layer (conv-5) of $\phi$ and $\psi$, and the network upto the attention blocks each producing $512$ feature maps of size $7 \times 7$. The attention blocks are followed by three new dense layers which project the attended feature maps onto the final latent space with dimensions $\mathbb{R}^{256}$. Besides, a spatial average pooling across the channels is applied on the outputs of $(\phi_\text{att}, \psi_\text{att})$ to obtain a single channel feature map of resolution $7 \times 7$. The encoder and decoder modules of $\mathcal{V}_{\alpha}$ and $\mathcal{V}_p$ are single dense layers of 128-d. The local and global domain classifiers $l(\cdot)$ and $f(\cdot)$, and the multi-class category classifier $h(\cdot)$ are also one dense layer each. For the semantic projection network $g(\cdot)$, $g_1$ and $g_3$ are three dense layers each. $g_2$ is a graph convolution layer, followed by the pooling and flattening layers. During training, VGG-16 layers prior to conv-5 are frozen while the proposed layers are updated.

\section{Results}
We choose the following state of the art methods, ZSIH~\citep{shen2018zero}, SEM-PCYC~\citep{dutta2019semantically}, Doodle2search~\citep{dey2019doodle},  Style-guide~\citep{dutta2019style}, AMD regularizer with SEM-PCYC and Style-guide~\citep{duttaadaptive}, respectively, for analyzing our performance. 
While SEM-PCYC, and Style-guide are based on adversarial training, ZS-SBIR utilizes variational encoder-decoder networks. AMD regularizer helps in tackling the data/class imbalance between the training and test sets. 
The performance of our full BDA-SketRet model on the Sketchy dataset with its two splits, the Tu-Berlin dataset and Quickdraw-extended datasets in comparison to the state of the art (SOTA) is reported in table~\ref{tab:comparison}.   {The mAP value for ZS-SBIR in Sketchy S2 is 43.5.} Similar to BDA-SketRet, these techniques report their performances using the VGG-16~\citep{simonyan2014very} based feature backbone networks. We have not compared our results with \citep{ijcai2020-137, CHAUDHURI2020104003} as the subset of test data considered is different from ours. For Sketchy, split 2 is considered to be more difficult than split 1 as it consists of test classes which are unseen to the ImageNet pre-trained networks. 

Amongst the other competing techniques, we find the inclusion of AMDreg boosts the performance of the baseline ZS-SBIR systems.  In spite of this, BDA-SketRet beats SEM-PCYC + AMD by a considerable margin of ~$3-5 \%$. The TU-Berlin dataset is challenging mainly due to the presence of class-wise as well as  domain-wise data imbalance. The performance of other competing techniques are extremely low. In contrast, we achieve a boost of ~$5 \%$ over the existing literature. The QuickDraw dataset is excessively large consisting of highly ambiguous sketches and is by far the most challenging dataset for ZS-SBIR. BDA-SketRet beats the P@100 value over the SOTA with a margin of ~$10\%$, while falls marginally to SEM-PCYC in the mAP value. 

Similar to ZS-SBIR, BDA-SketRet showcase overall improved performance measures for GZS-SBIR for all the datasets. In particular, BDA-SketRet produces high P@100 values of 41.3 for Sketchy (S1), 35.7 for TU-Berlin, and 28.6 for QuickDraw which are at least 2.5 \% more than the literary works. No prior approach report the GZS-SBIR score for S2 for Sketchy yet. However, we find that our performance in this case is substantially high. 

We note that the existing techniques only report P@100 or P@200 but not both. However, we feel that an effective ZS-SBIR system should produce high scores for both the metrics together. In this section, we report the performance of all the datasets, including the two splits of the Sketchy dataset with all the evaluation metrics in  table~\ref{tab:comparison2}. 
The top part report the ZS:SBIR results, while the bottom part reports the GZS:SBIR.
{
\setlength{\tabcolsep}{10pt}
\begin{table}\caption{Performance of the proposed BDA-SketRet with ZS-SBIR (top) and GZS-SBIR (bottom) on both the splits of Sketchy, TUB and Quickdraw datasets on all evaluation metrices.
}
\label{tab:comparison2}
\begin{center}
\resizebox{\linewidth}{!}{%
\begin{tabular}{lcccc}  \hline  
\textbf{Dataset} & \textbf{mAP}& \textbf{P@ 100}&\textbf{P@ 200}& \textbf{mAP @200}\\
    \hline  
Sketchy  (S 1) &43.7&51.4& 45.7&56.9 \\
Sketchy  (S 2) & 43.5  & 51.2 & 45.8 & 55.6 \\
TU-Berlin  & 37.4& 50.4& 43.8 & 54.4  \\
QuickDraw  & 15.4 & 44.0& 35.5&34.6 \\ \hline
Sketchy  (S 1) & 33.8 & 41.3& 31.4&38.6 \\
Sketchy  (S 2) &22.7 &25.1 &22.6 & 33.7 \\
TU-Berlin  &  25.1 & 35.7& 32.9& 33.3\\
QuickDraw  & 15.4 & 28.6& 29.5&27.4 \\   \hline  
\end{tabular}
}
\end{center}
\end{table}
}

%


\subsection{Evaluating Effect Of Input Modalities.} In the ZS-SBIR setup, the significance of the semantic information is imperative in maneuvering the alignment of the multi-modal data in the latent space. Different models yield different topological alignment of the classes in the latent space, which effectively causes the similar classes to cluster in a short range, while pushing apart the faraway classes. We consider the individual textual (300-d) and hierarchical embeddings as well as their concatenations and report the mAP values in table~\ref{tab:sideinfo} for both Sketchy and TU-Berlin. We observe that there is a variation of up to $4-6\%$ in the performance of BDA-SketRet by using different semantic information.  It is found that the individual semantic spaces provide superior performance than their pairwise combinations as neighborhood topology may not be consistent in different semantic spaces. 
We obtain the best performance of Sketchy is using the fasttext model, while it is Jiang-Conrath for TU-Berlin produces the best performance with a mAP of $37.4$.

 \setlength{\tabcolsep}{9pt}
\renewcommand{\arraystretch}{1}
\begin{table} {\caption{Effects of different semantic information on the mAP value for TU-Berlin and Sketchy (split 2) datasets.}\label{tab:sideinfo}%
}\centering
\resizebox{\columnwidth}{!}{
\scalebox{1}{
\begin{tabular}{cccccc} 
\textbf{W2v}& \textbf{Fasttext}& \textbf{Path} & \textbf{Jin-Con}& \textbf{Sketchy}  &\textbf{TUB}  \\
\hline
\rowcolor{Gray}\checkmark &  & & &41.1&37.4\\ 
 & \checkmark& & & \textbf{43.5}&36.5\\ 
\rowcolor{Gray} &  & \checkmark& & 40.2&32.9\\ 
 &  & &\checkmark & 41.9&\textbf{37.4}\\ 
\rowcolor{Gray} \checkmark &  &\checkmark& &40.1&31.4\\ 
 \checkmark &  &&\checkmark & 40.7&37.1\\ 
 \rowcolor{Gray} &\checkmark  &\checkmark& &39.1&32.8\\ 
   &\checkmark  &&\checkmark &39.7&34.6\\ \hline
\end{tabular}}
}
\end{table}

 \setlength{\tabcolsep}{1pt}
 \renewcommand{\arraystretch}{1}
\begin{table}\caption{Comparison of different visual backbones on BDA-SketRet and the corresponding SOTA. *~denotes mAP@200. SAKE uses a differently trained feature backbone.}\label{tab:pretrain}
\centering\resizebox{\columnwidth}{!}{
\scalebox{1}{
\begin{tabular}{@{\extracolsep{1pt}} l ccrcc }
 &\multicolumn{2}{c}{\textbf{BDA-SketRet}} &
\multicolumn{3}{c}{\textbf{State-of-the-Art}}\\
\cline{2-3}  \cline{4-6} 
\multirow{1}{*}{\textbf{Pretrain}} & \small{Sketchy}& TUB& Reference&\small{Sketchy}& TUB \\
\hline
VGG-16 &43.5 &37.4 &\multicolumn{3}{c}{ ---- Table~\ref{tab:comparison} ----} \\
ResNet-50 & 40.1& 33.2& SkechGCN \citep{zhang2020zero}& 38.2&32.4 \\
ResNet-152 &$43.0^*$ & $25.8^*$&SAN \citep{p2020stacked} &$24.0^*$ & $14.0^*$\\
SE-ResNet50 &$51.2^*$ & {41.0} &SAKE~\citep{liu2019semantic} &$49.7^*$ &47.5\\
\hline
\end{tabular}}}
\end{table}
\subsection{Evaluating Effect Of Feature Backbones.} Further, different backbone networks have been utilized by a few existing techniques for ZS-SBIR. It is unjust to directly compare them with the rest of the literary works which exploit the conventional VGG-16 framework. Hence, we deploy different encoder networks to train BDA-SketRet and compare with the respective approaches (table~\ref{tab:pretrain}) to provide a base-lining for the future endeavors. Similar to the semantic information, the chosen visual feature encoder affects the model performance considerably.
SkechGCN~\citep{zhang2020zero} considers the ResNet-50~\citep{he2016deep} architecture while SAN~\citep{p2020stacked} utilizes the ResNet 152~\citep{he2016deep}, both pre-trained on the Imagenet dataset. We also test the performance of our framework using the trending SE-ResNet-50 feature extractor.  {SAKE~\citep{liu2019semantic} uses a conditional-SE-ResNet 50~\citep{hu2018squeeze} architecture,  while using an auxiliary task to approximately map each image in the training set to the ImageNet semantic space. SE-ResNet is different from CSE-ResNet as it does not use any conditional variable. Similarly,  \citep{thong2020open} follow a different evaluation protocol from the remaining literature. While the other works use the entire seen classes along with the unseen classes for the GZS-SBIR experiments, in this paper the authors claim that they just use 20\% of the samples from the seen classes for evaluation. Hence, a direct comparison of results with~\citep{thong2020open,liu2019semantic} may not be fair. Apart from these two, overall it can be observed that BDA-SketRet beats the concerned techniques consistently when adopting the respective visual feature extractors.} 
%

\subsection{Ablation Studies And Qualitative Results}
%
\setlength{\tabcolsep}{9pt}
\renewcommand{\arraystretch}{0.9}
\begin{table}\caption{Ablation of loss functions and model components in terms of mAP for Sketchy (S2) and TU-Berlin. F denotes the full model. ($\mathcal{L}_\text{tri} = \mathcal{L}_\text{triplet}$, $\mathcal{L}_\text{sem} = \mathcal{L}_\text{semantic}$).}\label{tab:ablation}
\centering\resizebox{\columnwidth}{!}{
\scalebox{1}{
\begin{tabular}{m{0.3em}|lcc} 
&\textbf{Experimental set up} & \textbf{Sketchy} & \textbf{TUB}\\
\hline
\multirow{8}{2em}{\rotatebox{90}{\footnotesize{Losses} }}
&$\mathcal{L}_\text{sem} +  \mathcal{L}_\text{tri}$&27.6 & 20.1 \\
&$\mathcal{L}_\text{sem}  + \mathcal{L}_\text{tri} + \mathcal{L}_\text{t-skl}$&30.3 & 23.6 \\
&$\mathcal{L}_\text{sem}  + \mathcal{L}_\text{tri} + \mathcal{L}_\text{t-skl} + \mathcal{L}_\text{class} $&33.4 &  27.7\\
&$\mathcal{L}_\text{sem} + \mathcal{L}_\text{tri}  $ + $\mathcal{L}_\text{t-skl} + \mathcal{L}_\text{class} + \mathcal{L}_{2}$& 36.9& 33.5 \\
&$\mathcal{L}_\text{sem} + \mathcal{L}_\text{tri} $+ $\mathcal{L}_\text{t-skl} + \mathcal{L}_\text{class}$ + $\mathcal{L}_{3}$ & 40.1& 34.8 \\
&$\mathcal{L}_\text{sem} + \mathcal{L}_\text{tri} $+ $\mathcal{L}_{2}$ & 31.5& 29.2 \\
&$\mathcal{L}_\text{sem} + \mathcal{L}_\text{tri} $+ $\mathcal{L}_{3}$ & 33.2& 27.0\\
&$\mathcal{L}_\text{sem} + \mathcal{L}_\text{tri} $+ $\mathcal{L}_{2}$ + $\mathcal{L}_{3}$& 36.5& 32.9 \\
&$\mathcal{L}_\text{sem} + \mathcal{L}_\text{tri} $+ $\mathcal{L}_\text{t-skl} + \mathcal{L}_\text{class} + \mathcal{L}_{2}$ + $\mathcal{L}_{3}$ & \textbf{43.5}& \textbf{37.4} \\ \hline
\multirow{4}{2em}{\rotatebox{90}{\footnotesize{Model} }}
&$F$ w/o GCN & 41.2& 35.0  \\
&$F$ w/o attention block in local DA&38.6& 32.5 \\
&$F$ w/o local attention \& GCN & 34.2& 32.8 \\
\end{tabular}}}
\end{table}

\myparagraph{Ablation of model components:} The full model consists of a group of sub-modules, each contributing in its own way to enhance the performance. In the ablation analysis, the baseline network comprises of $\mathcal{L}_\text{triplet} + \mathcal{L}_\text{semantic}$. The global adaptation is performed on the latent features to reduce the domain-gap between the data from the two modalities by increasing the domain confusion. This improves the performance marginally, as seen from table~\ref{tab:ablation} .  This is expected to yield a class-wise overlapping embedding space for sketches and images.
Simply adding the binary domain classifier without the label classifier leads to mode collapse.  To avoid this and ensure class-wise discriminativeness, we add the full $\mathcal{L}_1$ loss and observe an increase in the overall performance. We then append the network with the local adaptation module applied on the intermediate feature maps to highlight important local constructs common to both the modalities.  When we add the cross-modal reconstruction modules, we observe significant improvements in the results (43.5 / 37.4).  {To study the individual contributions of $\mathcal{L}_2$ and $\mathcal{L}_3$, we use them individually and in conjunction with each other to the baseline model. A clear fall of about ~$7-10\%$ is observed, highlighting the contribution of the semantic adaptation module.} As evident from table~\ref{tab:ablation}, the full model incurs a boost of ~$12 -13 \%$ on the mAP values for both the datasets than the baseline, ranging from 27.6 to 43.5 in the Sketchy and 20.1 to 37.4 in the TU-Berlin. 

Further, we study the effects of the GCN module in $g(\cdot)$ and the spatial attention layers in $\phi_\text{att}$ and $\psi_\text{att}$, respectively. We observe a marginal performance drop of $1 -2\%$  when the GCN layer is removed from the full BDA-SketRet. Similarly, the attention module is crucial in highlighting the domain-invariant mid-level features and BDA-SketRet without the attention layers is found to marginally degrade the performance.  We also look into the effect of removing the GCN module for the GZS-SBIR experiments and notice a drop of ~$10\%$ in mAP.

\begin{figure}\centering
 \subfloat[\footnotesize{Sample sketch and photo images.}]{\includegraphics[width=0.2\linewidth]{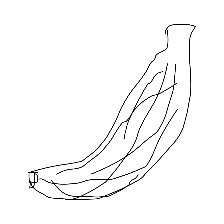}  \includegraphics[width=0.2\linewidth]{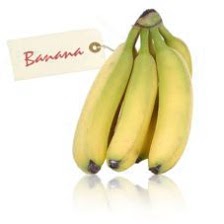} 
 \includegraphics[width=0.2\linewidth]{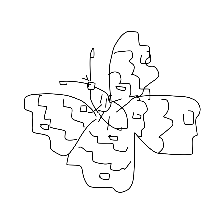} 
 \includegraphics[width=0.2\linewidth]{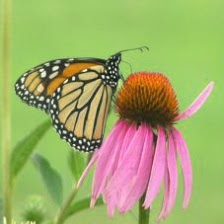}}\hfill 
 \subfloat[\footnotesize{Grad-CAM plots highlighting the ROI on the model without  $\mathcal{L}_\text{dom}^\text{global}$, $\mathcal{L}_\text{t-skl}$, $\mathcal{L}_2$ and $\mathcal{L}_3$.}]{\includegraphics[width=0.2\linewidth]{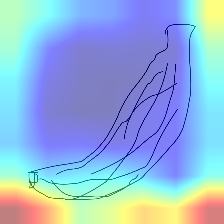}
 \includegraphics[width=0.2\linewidth]{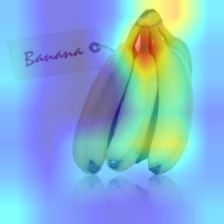}
  \includegraphics[width=0.2\linewidth]{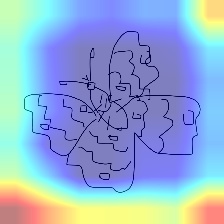}
 \includegraphics[width=0.2\linewidth]{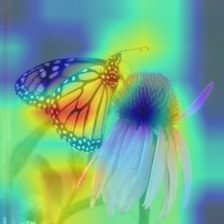}}\vspace{-2mm}\\
 \subfloat[\footnotesize{Grad-CAM plots highlighting the ROI on the model without $\mathcal{L}_\text{t-skl}$, $\mathcal{L}_2$ and $\mathcal{L}_3$ losses.}]{\includegraphics[width=0.2\linewidth]{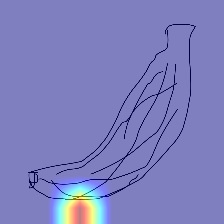}
 \includegraphics[width=0.2\linewidth]{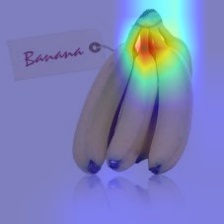}
  \includegraphics[width=0.2\linewidth]{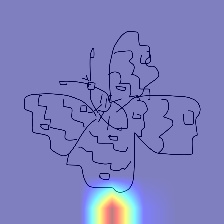}
 \includegraphics[width=0.2\linewidth]{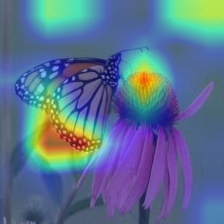}}\hfill
 \subfloat[\footnotesize{Grad-CAM plots highlighting the ROI on the model without $\mathcal{L}_2$ and $\mathcal{L}_3$ losses.}]{\includegraphics[width=0.2\linewidth]{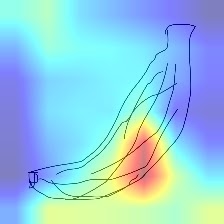}
           \includegraphics[width=0.2\linewidth]{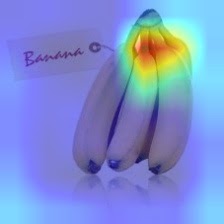}
           \includegraphics[width=0.2\linewidth]{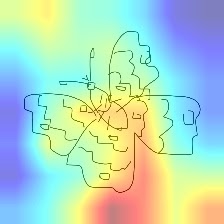}
          \includegraphics[width=0.2\linewidth]{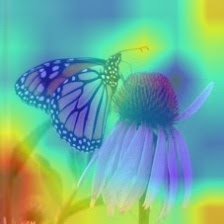}}\vspace{-2mm}\\
 \subfloat[\footnotesize{Grad-CAM plots highlighting the ROI when trained with the full model.}]{\includegraphics[width=0.2\linewidth]{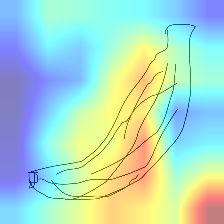}
          \includegraphics[width=0.2\linewidth]{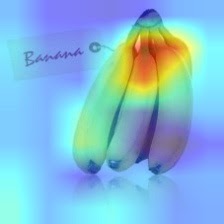}
          \includegraphics[width=0.2\linewidth]{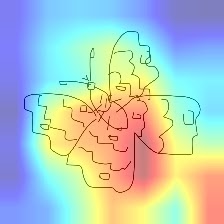}
          \includegraphics[width=0.2\linewidth]{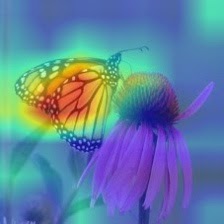}}\vspace{-3mm}\\
\caption{Ablation of domain adaptation stages using Grad-CAM plots to highlight the region of interest (ROI).}
\label{fig:gradcam}
\end{figure}

\begin{figure*}
\addtolength{\tabcolsep}{-8pt} 
 \scalebox{1}{\begin{tabular}{ccccccccc c| ccccccccc}
   {\includegraphics[width=9mm, height = 9mm]{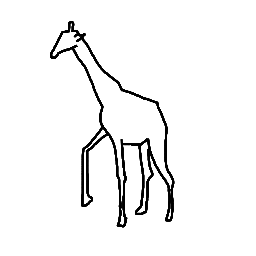}}&
   {\includegraphics[width=9mm, height = 9mm]{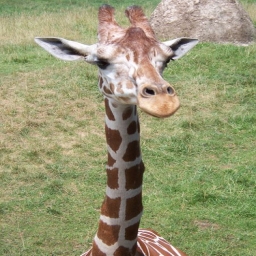}}&
   {\includegraphics[width=9mm, height = 9mm]{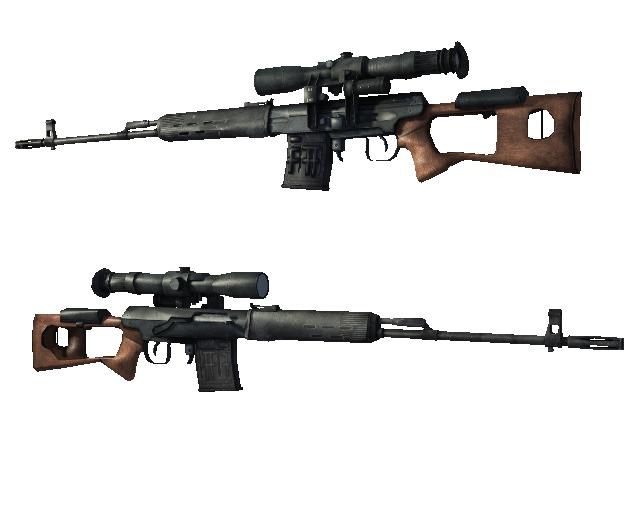}}&
   {\includegraphics[width=9mm, height = 9mm]{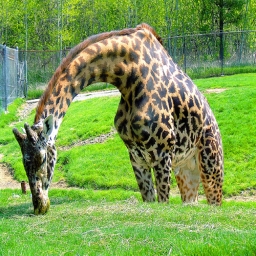}}&
   {\includegraphics[width=9mm, height = 9mm]{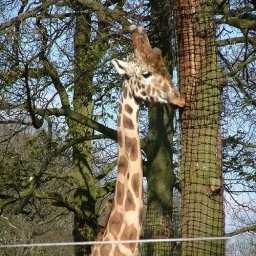}}&
   {\includegraphics[width=9mm, height = 9mm]{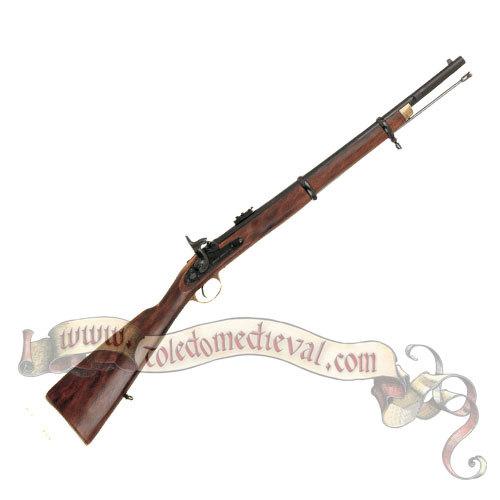}}&
   {\includegraphics[width=9mm, height = 9mm]{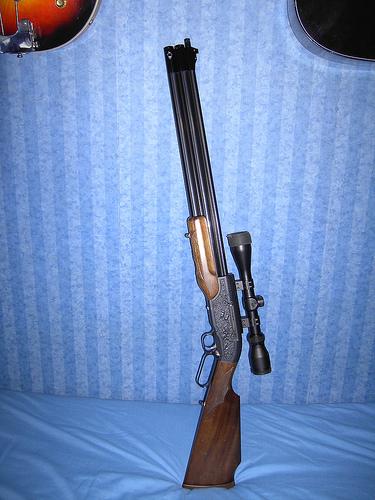}}&
   {\includegraphics[width=9mm, height = 9mm]{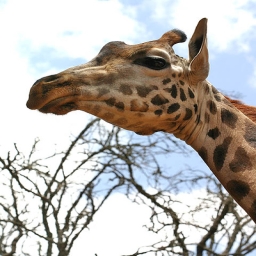}}&
   {\includegraphics[width=9mm, height = 9mm]{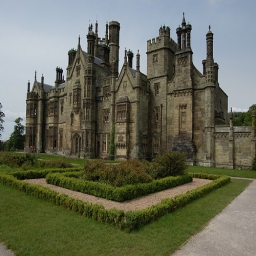}}&
\hspace{5pt}&\hspace{5pt}
   {\includegraphics[width=9mm, height = 9mm]{fig/giraffe_s.png}}&
   {\includegraphics[width=9mm, height = 9mm]{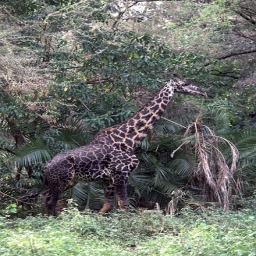}}&
   {\includegraphics[width=9mm, height = 9mm]{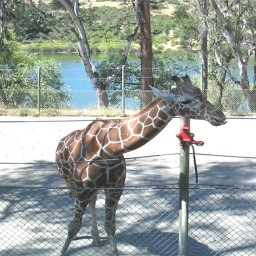}}&
   {\includegraphics[width=9mm, height = 9mm]{fig/giraffe_7.jpg}}&
   {\includegraphics[width=9mm, height = 9mm]{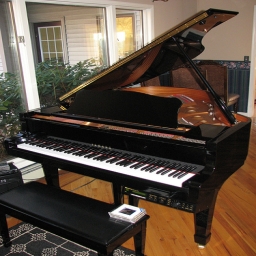}}&
   {\includegraphics[width=9mm, height = 9mm]{fig/giraffe_5.jpg}}&
   {\includegraphics[width=9mm, height = 9mm]{fig/piano_1.jpg}}&
   {\includegraphics[width=9mm, height = 9mm]{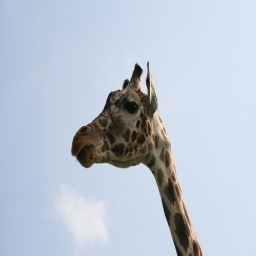}}&
   {\includegraphics[width=9mm, height = 9mm]{fig/giraffe_3.jpg}}\\
   
        \scriptsize{\tt Giraffe}& \greencheck & \bluestar &\greencheck &\greencheck &\bluestar &\bluestar &\greencheck &\redcross  & 
        
     &\scriptsize{\tt Giraffe}& \greencheck & \greencheck &\greencheck &\redcross &\greencheck &\redcross &\greencheck &\greencheck  \\

   {\includegraphics[width=9mm, height = 9mm]{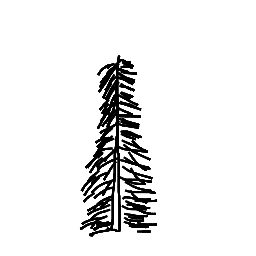}}&
   {\includegraphics[width=9mm, height = 9mm]{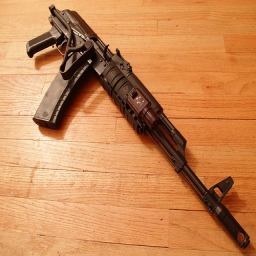}}&
   {\includegraphics[width=9mm, height = 9mm]{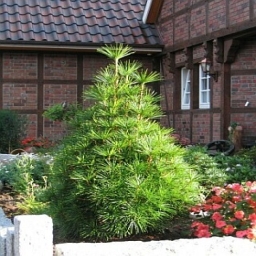}}&
   {\includegraphics[width=9mm, height = 9mm]{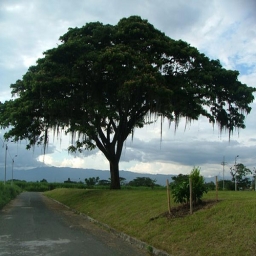}}&
   {\includegraphics[width=9mm, height = 9mm]{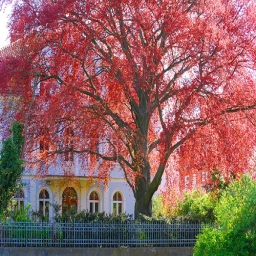}}&
   {\includegraphics[width=9mm, height = 9mm]{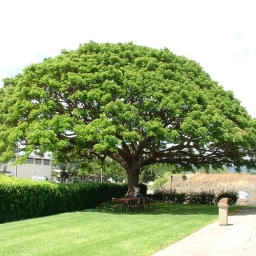}}&
   {\includegraphics[width=9mm, height = 9mm]{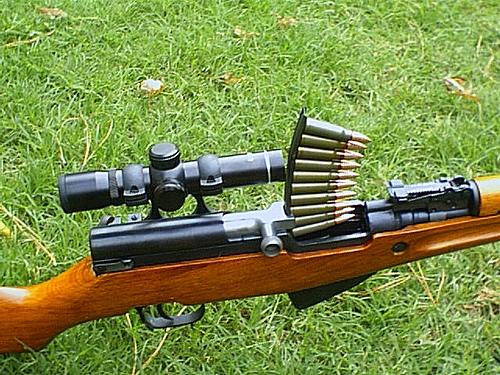}}&
   {\includegraphics[width=9mm, height = 9mm]{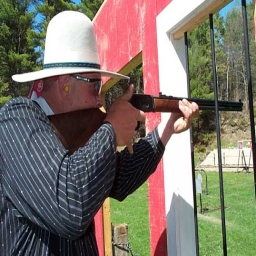}}&
   {\includegraphics[width=9mm, height = 9mm]{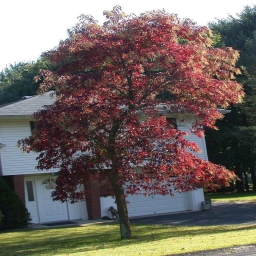}}&
   \hspace{5pt} & \hspace{5pt}
   {\includegraphics[width=9mm, height = 9mm]{fig/tree_s.png}}&
   {\includegraphics[width=9mm, height = 9mm]{fig/tree_4.jpg}}&
   {\includegraphics[width=9mm, height = 9mm]{fig/tree_6.jpg}}&
   {\includegraphics[width=9mm, height = 9mm]{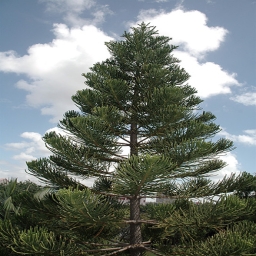}}&
   {\includegraphics[width=9mm, height = 9mm]{fig/tree_7.jpg}}&
   {\includegraphics[width=9mm, height = 9mm]{fig/tree_5.jpg}}&
   {\includegraphics[width=9mm, height = 9mm]{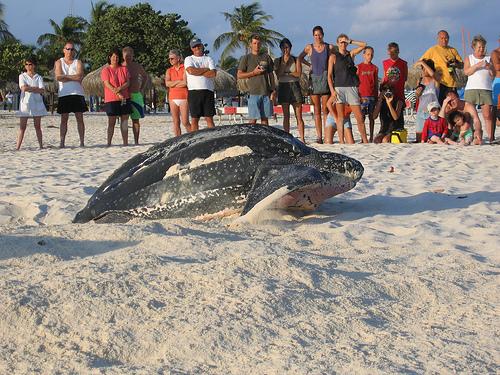}}&
   {\includegraphics[width=9mm, height = 9mm]{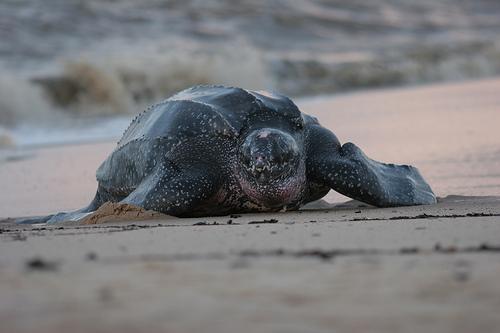}}&
   {\includegraphics[width=9mm, height = 9mm]{fig/tree_5.jpg}}\\  
   \scriptsize{\tt Tree}& \bluestar & \greencheck &\greencheck &\greencheck &\greencheck &\bluestar &\bluestar &\greencheck &
   
   & \scriptsize{\tt Tree}& \greencheck & \greencheck &\greencheck &\greencheck &\greencheck &\redcross &\redcross &\greencheck  \\
   
   {\includegraphics[width=9mm, height = 9mm]{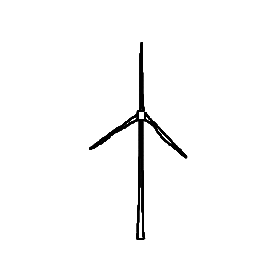}}&
   {\includegraphics[width=9mm, height = 9mm]{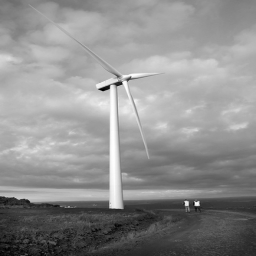}}&
   {\includegraphics[width=9mm, height = 9mm]{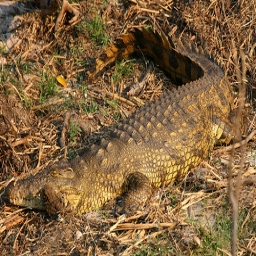}}&
   {\includegraphics[width=9mm, height = 9mm]{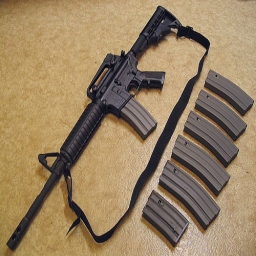}}&
   {\includegraphics[width=9mm, height = 9mm]{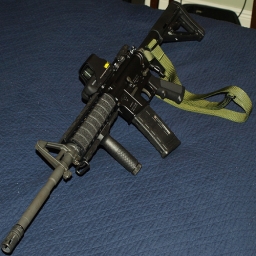}}&
   {\includegraphics[width=9mm, height = 9mm]{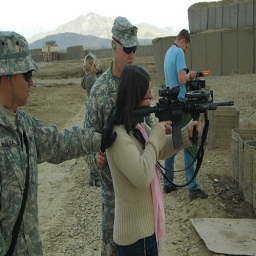}}&
   {\includegraphics[width=9mm, height = 9mm]{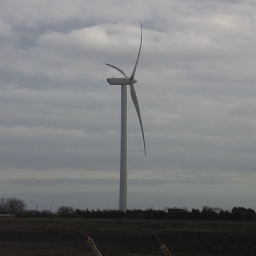}}&
   {\includegraphics[width=9mm, height = 9mm]{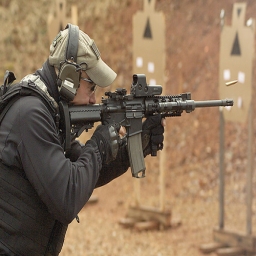}}&
   {\includegraphics[width=9mm, height = 9mm]{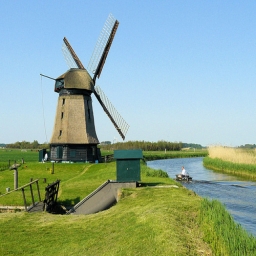}}&
     \hspace{5pt}&\hspace{5pt}
   {\includegraphics[width=9mm, height = 9mm]{fig/windmill_s.png}}&
   {\includegraphics[width=9mm, height = 9mm]{fig/windmill_8.jpg}}&
   {\includegraphics[width=9mm, height = 9mm]{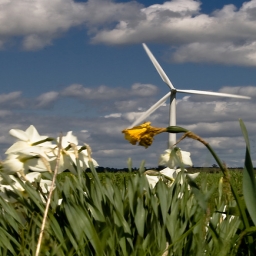}}&
   {\includegraphics[width=9mm, height = 9mm]{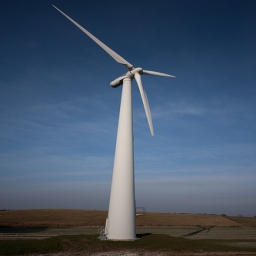}}&
   {\includegraphics[width=9mm, height = 9mm]{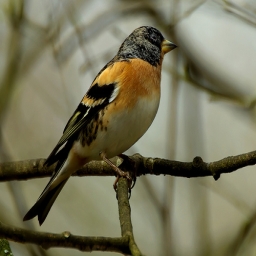}}&
   {\includegraphics[width=9mm, height = 9mm]{fig/windmill_6.jpg}}&
   {\includegraphics[width=9mm, height = 9mm]{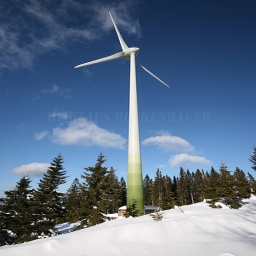}}&
   {\includegraphics[width=9mm, height = 9mm]{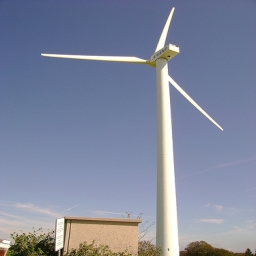}}&
   {\includegraphics[width=9mm, height = 9mm]{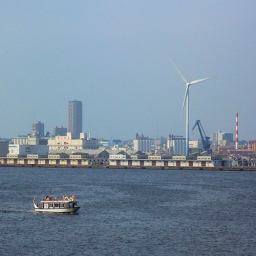}}\\

   \scriptsize{\tt Windmill}& \greencheck & \redcross &\bluestar &\bluestar &\bluestar &\greencheck &\bluestar &\greencheck &
    &~\scriptsize{\tt Windmill}& \greencheck & \greencheck &\greencheck &\redcross &\greencheck &\greencheck &\greencheck &\greencheck  \\  
  \end{tabular}}\addtolength{\tabcolsep}{6pt}  
    \caption{Top-8 retrieval instances for a few sketch queries from the Sketchy dataset  {without using the local adaptation} on the left and BDA-SketRet on the right. The green checks denote correctly retrieved classes, while the red crosses denote incorrect class images. The blue stars denote the hub instances.
}
\label{fig:hubs} 
\end{figure*}

\myparagraph{Qualitative analysis of hubness.}   {{ Hubness} problem~\citep{radovanovic2010hubs} occurs when a model has a training bias and retrieves images only from a subset of the available categories. This occurs as some embedding vectors of images (also called as ``hubs'') appear in the nearest neighborhood of many test query sketches. An ineffective feature alignment between the visual modalities may trigger the generation of hubs. In order to show the role of domain losses in retrieving more discriminative image samples, we first train the model with just the adversarial global adaptation (without $\mathcal{L}_2$ and $\mathcal{L}_3$), followed by training the entire BDA-SketRet model. In the first case, we notice the presence of hubs. Precisely, the left column of Fig.~\ref{fig:hubs} shows a scenario where the instances of {\tt rifle} class are retrieved for nearly many query samples as its embeddings are cluttered in the feature space with multiple classes.} This adversely affects the overall performance. In the second case, no particular class is visibly found to clutter the retrieval results in the latent space. It depicts that by jointly using the global and local adaptations and using cross-modal reconstruction modules, we achieve a hub-free retrieval results for the same set of queries. 

\begin{figure}
    \centering
\subfloat[\footnotesize{Original word embeddings.}]{\includegraphics[width = 0.32\linewidth]{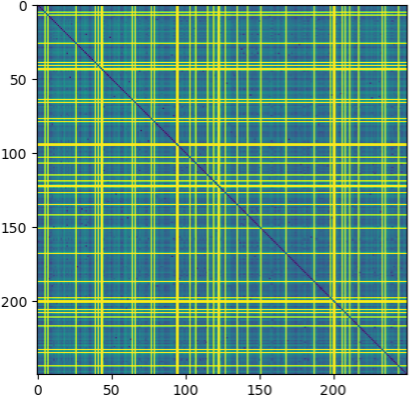}}\hspace{0.5mm}
 \subfloat[\footnotesize{Auto-encoded word embeddings in latent space.}]{\includegraphics[width = 0.32\linewidth]{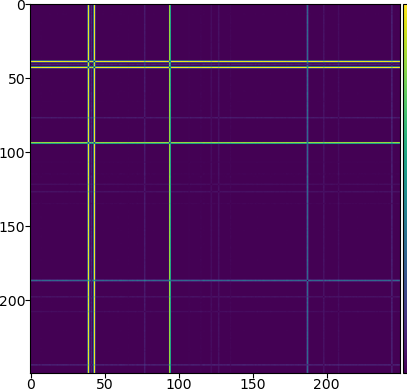}}\hspace{0.5mm}
\subfloat[\footnotesize{GCN preserved word embeddings in latent space.}]{\includegraphics[width = 0.32\linewidth]{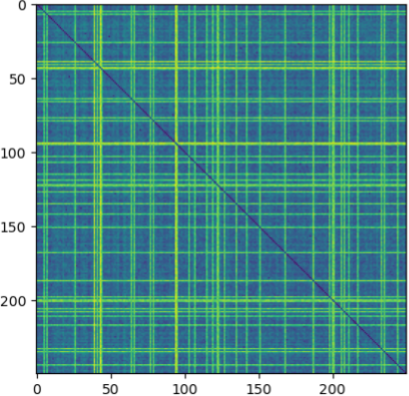}}
\caption{Co-variance matrices of the word embeddings under different experimental set-ups.}
    \label{fig:semantic}
\end{figure}

\myparagraph{Preservation of semantic space using GCN}
In this section, we aim to show that the leveraging a graph convolution network (GCN), we preserve the semantic topology in the embedding feature space. The semantic information has an important role to play for attaining an improved ZS-SBIR inference. Generally, the visual modalities are aligned to a fixed semantic space. While \citep{dutta2019semantically} uses an encoded semantic space obtained via a semantic auto-encoder, methods of \citep{dey2019doodle, dutta2019style} reconstruct the original semantic vectors from the visual information. These methods could lead to a drastic change of the semantic topology  in the embedding feature space. Fig.~\ref{fig:semantic} (a) from the main manuscript shows the co-variance matrix of the 250 classes of the TU-Berlin dataset, while Fig.~\ref{fig:semantic} (b) shows the co-variance matrix of the same after passing through a semantic auto-encoder. It can be seen that the semantic topology is visibly distorted, which would lead to improper mapping of visual and semantic space for the unseen classes.

In the proposed BDA-SketRet framework, we utilize graph CNN \citep{kipf2016semi} to encode the structural similarity among the different classes. We model an independent latent space given the original modalities but constraint the space to be influenced by the characteristics of the original semantic space. Fig.~\ref{fig:semantic} (c) shows the co-variance matrix of the 250 classes of the TU-Berlin dataset. It can be seen that the semantic topology is visibly preserved in the embedding space. In effect, this helps us in obtaining a better mapping of the visual and semantic space for the unseen classes.


\subsection{Grad-CAM Visualization} Gradient-weighted class activation mapping~\citep{selvaraju2017grad} (Grad-CAM) primarily uses the gradients of the target class at the final convolution layer to synthesize an intermediate localization map which highlights the most important regions in the image. It effectively helps in displaying the region which gets the most importance for any particular target-class.

\begin{figure*}
\addtolength{\tabcolsep}{-8.5pt} 
    \begin{center}
    \scalebox{0.99}{\begin{tabular}{cccccccccc}
\includegraphics[width=0.1\linewidth]{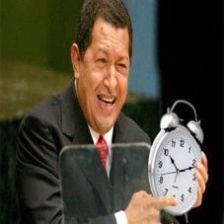}&
 \includegraphics[width=0.1\linewidth]{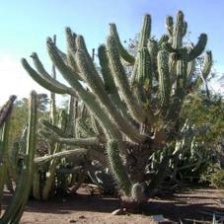}&
 \includegraphics[width=0.1\linewidth]{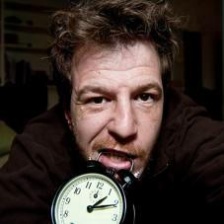}&
 \includegraphics[width=0.1\linewidth]{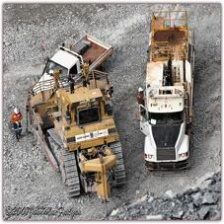}&
 \includegraphics[width=0.1\linewidth]{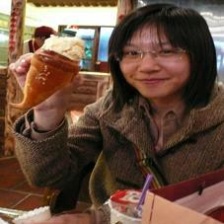}&
 \includegraphics[width=0.1\linewidth]{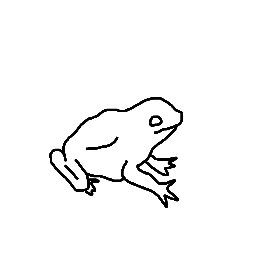} &
\includegraphics[width=0.1\linewidth]{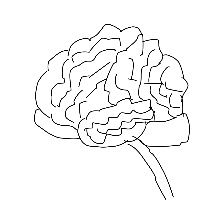} &
\includegraphics[width=0.1\linewidth]{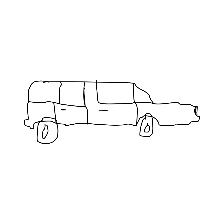}&
\includegraphics[width=0.1\linewidth]{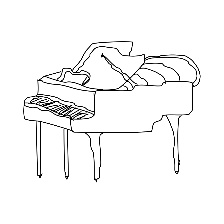}&
\includegraphics[width=0.1\linewidth]{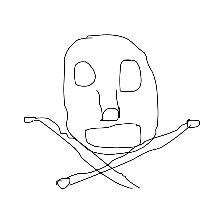}\\
 \subfloat[\scriptsize{Alarmclock}]{\includegraphics[width=0.1\linewidth]{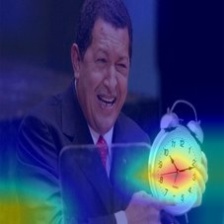}}&
 \subfloat[\footnotesize{Cactus}]{\includegraphics[width=0.1\linewidth]{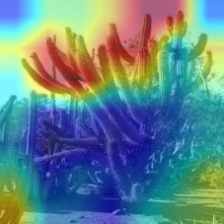}}&
 \subfloat[\footnotesize{Clock}]{\includegraphics[width=0.1\linewidth]{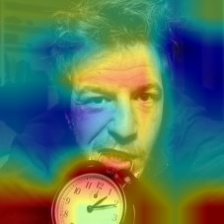}}&
 \subfloat[\footnotesize{Bulldozer}]{\includegraphics[width=0.1\linewidth]{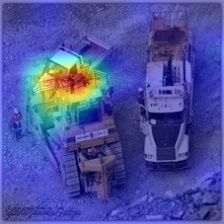}}&
 \subfloat[\footnotesize{Ice-cream}]{\includegraphics[width=0.1\linewidth]{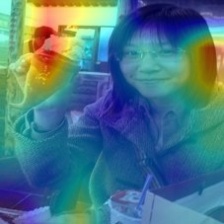}}&
\subfloat[\footnotesize{Frog}]{\includegraphics[width=0.1\linewidth]{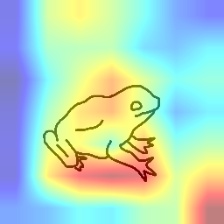}} &
\subfloat[\footnotesize{Brain}]{\includegraphics[width=0.1\linewidth]{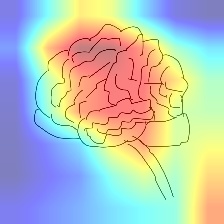}} &
\subfloat[\footnotesize{SUV}]{\includegraphics[width=0.1\linewidth]{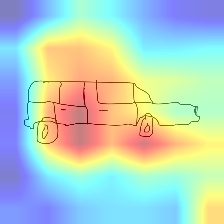}}&
\subfloat[\footnotesize{Piano}]{\includegraphics[width=0.1\linewidth]{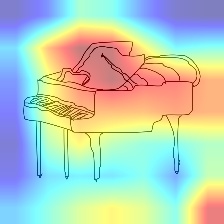}}&
\subfloat[\footnotesize{Skull}]{\includegraphics[width=0.1\linewidth]{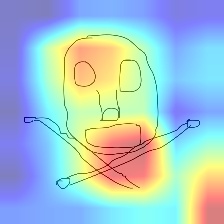}}\\
    \end{tabular}}
        \end{center}
    \caption{Grad-CAM plots of full model of a few sample photo and sketch images.}
    \label{fig:photogradcam}
\end{figure*}

\myparagraph{Effect of bi-level adaptation.} We ablate among the various domain adaptation modules in the network and provide a visualization of the same in Fig.~\ref{fig:gradcam} and  show the Grad-CAM plots of both the models for a few sketch and photo images.  

In Fig.~\ref{fig:gradcam} (b), we train the model without the $\mathcal{L}_\text{dom}^\text{global}$, $\mathcal{L}_\text{t-skl}$, $\mathcal{L}_2$ and $\mathcal{L}_3$ losses. Notice that this causes improper distribution of weights especially in the sketch images. The network primarily learns from the background of the sketch images, while for the photo images, the region of importance is scattered all over the image. Adding the $\mathcal{L}_\text{dom}^\text{global}$, relatively constricts the scattered region of importance to a more localized form. When we further go ahead and add the $\mathcal{L}_\text{t-skl}$ loss to the model, from Fig.~\ref{fig:gradcam} (d) we start noticing that the alignment of important region in the sketch images start improving. We can see that the full model produces better highlight to the local constructs. 

This establishes our claim that the notion of fine-grained domain adaptation helps in obtaining a more discriminative latent space to combat hubness and negative knowledge transfer judiciously. The Grad-CAM plots displays the region which gets the most importance (ROI) when we train the model. We can see that the full model produces better highlight to the local constructs.  {Notice in the butterfly image, without the bi-level adaptation, the attention is splattered across the image and this leads to incorrect class assignment during retrieval. Using the full model of BDA-SketRet with bi-level adaptation, although the foreground consists of both the butterfly and the flower, the importance is layed properly on the concerned {\tt butterfly} class. }

\myparagraph{Qualitative results.} In Fig.~\ref{fig:photogradcam} we show the GradCAM plots of a few sample photo and sketch images. From the photo images it can be seen that query class is properly highlighted in each image, keeping the remaining image as a background. For example, it can be seen from the {\tt Alarmclock} images that the attended region is carefully just the clock, leaving out the human as the background. Similarly, the dendritic-structure of the {\tt cactus} receives the most attention as they contribute more towards the overall recognition of its class. It can also be seen that out of the two power vehicles, the network correctly chooses the {\tt bulldozer} and puts more weight on it.

\begin{figure*}\center
   {\includegraphics[width=\linewidth]{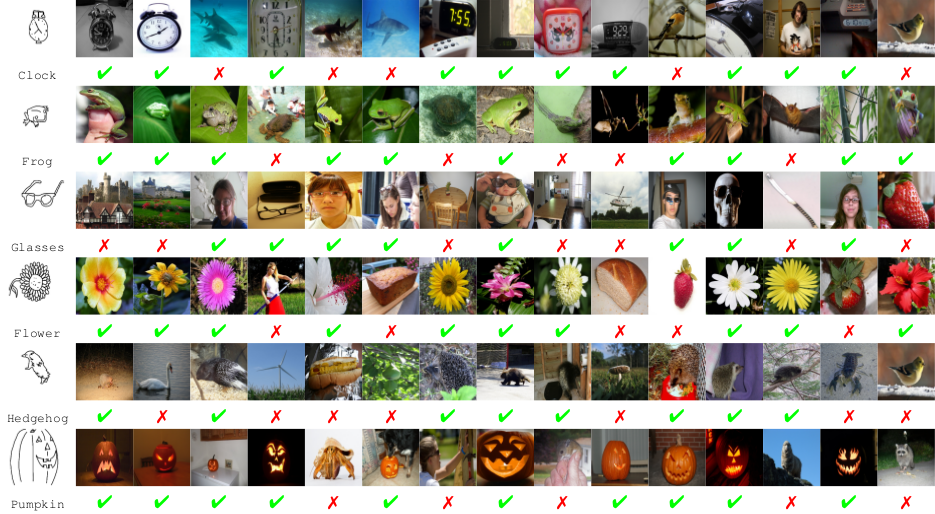}}
\caption{Top-15 retrieval instances for a few sketch queries from the Sketchy dataset using the full model. The green checks denote correctly retrieved classes, while the red crosses denote images from incorrect class.  Notice that there are no hub instances generated here.}
\label{fig:zssbir} 
\end{figure*}

Similarly, for the sketch classes it is seen that the highlighted region is the most important characteristic part of the sketch. For example, for {\tt frog}, the webbed feet are the most important region, while for a {\tt skull}, the cross-bones commonly drawn under skulls along with the hollow eye sockets conveyed the most information. It can be noted that since CNNs have a tendency to learn from image textures~\citep{geirhos2018imagenet}, the network provides more weights towards the image boundaries (as seen from Fig.~\ref{fig:gradcam} (b)). This is exactly the reason why sketch-based learning tasks are challenging. The proposed BDA-SketRet helps the network to learn the most important regions and their corresponding weights properly even from sketch images as shown in Fig.~\ref{fig:photogradcam}. In Fig.~\ref{fig:zssbir}, we show the ZS-SBIR results for a few sample sketches from the Sketchy-extended dataset.
\begin{table}\caption{Sensitivity analysis of hyper-parameters in terms of mAP value for TU-Berlin and Sketchy (split 2) datasets.}\label{tab:crossval}
\centering
\resizebox{\columnwidth}{!}{
\scalebox{1}{
\addtolength{\tabcolsep}{8pt} 
\begin{tabular}{ccccc} 
 $\mu$& $\lambda$ & $\beta$ & \textbf{Sketchy}  &\textbf{TUB}  \\
\hline
\rowcolor{Gray}1 & 1 & 0.0001& 41.8&\textbf{37.5}\\ 
0.1 & 1& 0.0001& 42.2& 37.1\\
\rowcolor{Gray}0.01 & 1 &0.0001 & 42.7& 36.7\\ 
0.1 & 0.1 & 0.0001&\textbf{43.5}& 35.2\\ 
 \rowcolor{Gray}0.1 & 0.01 &0.0001 & 40.6 &32.0\\ 
  1& 1 & 0.001 & 42.5& 35.9\\ 
 \rowcolor{Gray}1 & 1 &  0.01& 38.5&33.4\\
   1& 1  & 0.00001& 41.6 &36.8\\
\end{tabular}}}
\end{table}

\subsection{Sensitivity To Hyper-parameters}
{In the proposed work, we have primarily 3 hyper-parameters, i.e., $\lambda$, $\beta$, and $\mu$ from the equations 1 and 2 of the main manuscript. For tuning the hyper-parameters, we follow the standard protocol of the zero-shot learning community by splitting the training data for cross-validation into training and validation data (psuedo-seen and psudeo-unseen split). The overall performance of the proposed framework by varying these parameters with some of the settings are shown in Table~\ref{tab:crossval} for both Sketchy and TU-Berlin in terms of mAP values. We experiment the model performance by choosing different values in the range of 0.01 to 1 for $\mu$ and $\lambda$, while 0.00001 to 0.01 for $\beta$, as the $\mathcal{L}_\text{t-skl}$ loss is found to converge the fastest amongst the rest. We choose the combination of $\mu$, $\lambda$, and $\beta$ values that result in the best model performance. }

\section{Conclusions}
We introduce a novel ZS-SBIR framework called BDA-SketRet in this paper. The main premise of our model is to perform improved alignment between the image and sketch features based on both the mid-level and high-level CNN based feature embeddings. Together, we introduce two generative cross-modal reconstruction modules to ensure the learning of robust modality-independent features. We further propose to project the semantic information into the shared latent space through a two-stream fusion network by jointly exploiting both the prototypes and the semantic class neighborhood. Overall, BDA-SketRet learns a discriminative and compact latent space and wisely tackles both the negative transfer and the hubness issues of domain adaptation and ZSL, respectively. 
Experimentally, we outperform the recent techniques in all the performance metrics on all the existing datasets. We are currently interested in extending BDA-SketRet to support the paradigm of lifelong learning.


\bibliography{mybib}

\begin{thebibliography}{53}
\expandafter\ifx\csname natexlab\endcsname\relax\def\natexlab#1{#1}\fi
\providecommand{\url}[1]{\texttt{#1}}
\providecommand{\href}[2]{#2}
\providecommand{\path}[1]{#1}
\providecommand{\DOIprefix}{doi:}
\providecommand{\ArXivprefix}{arXiv:}
\providecommand{\URLprefix}{URL: }
\providecommand{\Pubmedprefix}{pmid:}
\providecommand{\doi}[1]{\href{http://dx.doi.org/#1}{\path{#1}}}
\providecommand{\Pubmed}[1]{\href{pmid:#1}{\path{#1}}}
\providecommand{\bibinfo}[2]{#2}
\ifx\xfnm\relax \def\xfnm[#1]{\unskip,\space#1}\fi
\bibitem[{Alemi et~al.(2017)Alemi, Fischer, Dillon and Murphy}]{45903}
\bibinfo{author}{Alemi, A.}, \bibinfo{author}{Fischer, I.},
  \bibinfo{author}{Dillon, J.}, \bibinfo{author}{Murphy, K.},
  \bibinfo{year}{2017}.
\newblock \bibinfo{title}{Deep variational information bottleneck}, in:
  \bibinfo{booktitle}{ICLR}.
\bibitem[{Bay et~al.(2006)Bay, Tuytelaars and Van~Gool}]{bay2006surf}
\bibinfo{author}{Bay, H.}, \bibinfo{author}{Tuytelaars, T.},
  \bibinfo{author}{Van~Gool, L.}, \bibinfo{year}{2006}.
\newblock \bibinfo{title}{Surf: Speeded up robust features}, in:
  \bibinfo{booktitle}{ECCV}.
\bibitem[{Ben-David et~al.(2009)Ben-David, Blitzer, Crammer, Kulesza, Pereira
  and Vaughan}]{BenDavid2009ATO}
\bibinfo{author}{Ben-David, S.}, \bibinfo{author}{Blitzer, J.},
  \bibinfo{author}{Crammer, K.}, \bibinfo{author}{Kulesza, A.},
  \bibinfo{author}{Pereira, F.C.}, \bibinfo{author}{Vaughan, J.W.},
  \bibinfo{year}{2009}.
\newblock \bibinfo{title}{A theory of learning from different domains}.
\newblock \bibinfo{journal}{Machine Learning} \bibinfo{volume}{79},
  \bibinfo{pages}{151--175}.
\bibitem[{Bojanowski et~al.(2017)Bojanowski, Grave, Joulin and
  Mikolov}]{bojanowski2017enriching}
\bibinfo{author}{Bojanowski, P.}, \bibinfo{author}{Grave, E.},
  \bibinfo{author}{Joulin, A.}, \bibinfo{author}{Mikolov, T.},
  \bibinfo{year}{2017}.
\newblock \bibinfo{title}{Enriching word vectors with subword information}.
\newblock \bibinfo{journal}{ACL} .
\bibitem[{Bui et~al.(2017)Bui, Ribeiro, Ponti and Collomosse}]{bui2017compact}
\bibinfo{author}{Bui, T.}, \bibinfo{author}{Ribeiro, L.},
  \bibinfo{author}{Ponti, M.}, \bibinfo{author}{Collomosse, J.},
  \bibinfo{year}{2017}.
\newblock \bibinfo{title}{Compact descriptors for sketch-based image retrieval
  using a triplet loss convolutional neural network}.
\newblock \bibinfo{journal}{CVIU} .
\bibitem[{Chaudhuri et~al.(2020)Chaudhuri, Banerjee, Bhattacharya and
  Datcu}]{CHAUDHURI2020104003}
\bibinfo{author}{Chaudhuri, U.}, \bibinfo{author}{Banerjee, B.},
  \bibinfo{author}{Bhattacharya, A.}, \bibinfo{author}{Datcu, M.},
  \bibinfo{year}{2020}.
\newblock \bibinfo{title}{Crossatnet - a novel cross-attention based framework
  for sketch-based image retrieval}.
\newblock \bibinfo{journal}{IVC} .
\bibitem[{Chopra et~al.(2005)Chopra, Hadsell and LeCun}]{chopra2005learning}
\bibinfo{author}{Chopra, S.}, \bibinfo{author}{Hadsell, R.},
  \bibinfo{author}{LeCun, Y.}, \bibinfo{year}{2005}.
\newblock \bibinfo{title}{Learning a similarity metric discriminatively, with
  application to face verification}, in: \bibinfo{booktitle}{CVPR}.
\bibitem[{Dalal and Triggs(2005)}]{dalal2005histograms}
\bibinfo{author}{Dalal, N.}, \bibinfo{author}{Triggs, B.},
  \bibinfo{year}{2005}.
\newblock \bibinfo{title}{Histograms of oriented gradients for human
  detection}, in: \bibinfo{booktitle}{CVPR}.
\bibitem[{Deng et~al.(2009)Deng, Dong, Socher, Li, Li and
  Fei-Fei}]{deng2009imagenet}
\bibinfo{author}{Deng, J.}, \bibinfo{author}{Dong, W.},
  \bibinfo{author}{Socher, R.}, \bibinfo{author}{Li, L.J.},
  \bibinfo{author}{Li, K.}, \bibinfo{author}{Fei-Fei, L.},
  \bibinfo{year}{2009}.
\newblock \bibinfo{title}{Imagenet: A large-scale hierarchical image database},
  in: \bibinfo{booktitle}{CVPR}.
\bibitem[{Dey et~al.(2019)Dey, Riba, Dutta, Llados and Song}]{dey2019doodle}
\bibinfo{author}{Dey, S.}, \bibinfo{author}{Riba, P.}, \bibinfo{author}{Dutta,
  A.}, \bibinfo{author}{Llados, J.}, \bibinfo{author}{Song, Y.Z.},
  \bibinfo{year}{2019}.
\newblock \bibinfo{title}{Doodle to search: Practical zero-shot sketch-based
  image retrieval}, in: \bibinfo{booktitle}{CVPR}.
\bibitem[{Dutta and Akata(2019)}]{dutta2019semantically}
\bibinfo{author}{Dutta, A.}, \bibinfo{author}{Akata, Z.}, \bibinfo{year}{2019}.
\newblock \bibinfo{title}{Semantically tied paired cycle consistency for
  zero-shot sketch-based image retrieval}, in: \bibinfo{booktitle}{CVPR}.
\bibitem[{Dutta and Akata(2020)}]{dutta2020semantically}
\bibinfo{author}{Dutta, A.}, \bibinfo{author}{Akata, Z.}, \bibinfo{year}{2020}.
\newblock \bibinfo{title}{Semantically tied paired cycle consistency for
  any-shot sketch-based image retrieval}.
\newblock \bibinfo{journal}{IJCV} .
\bibitem[{Dutta and Biswas(2019)}]{dutta2019style}
\bibinfo{author}{Dutta, T.}, \bibinfo{author}{Biswas, S.},
  \bibinfo{year}{2019}.
\newblock \bibinfo{title}{Style-guided zero-shot sketch-based image
  retrieval.}, in: \bibinfo{booktitle}{BMVC}.
\bibitem[{Dutta et~al.(2020)Dutta, Singh and Biswas}]{duttaadaptive}
\bibinfo{author}{Dutta, T.}, \bibinfo{author}{Singh, A.},
  \bibinfo{author}{Biswas, S.}, \bibinfo{year}{2020}.
\newblock \bibinfo{title}{Adaptive margin diversity regularizer for handling
  data imbalance in zero-shot sbir}, in: \bibinfo{booktitle}{ECCV}.
\bibitem[{{Dutta} et~al.(2020){Dutta}, {Singh} and
  {Biswas}}]{dutta2020styleguide}
\bibinfo{author}{{Dutta}, T.}, \bibinfo{author}{{Singh}, A.},
  \bibinfo{author}{{Biswas}, S.}, \bibinfo{year}{2020}.
\newblock \bibinfo{title}{Styleguide: Zero-shot sketch-based image retrieval
  using style-guided image generation}.
\newblock \bibinfo{journal}{IEEE TM} .
\bibitem[{Eitz et~al.(2012)Eitz, Hays and Alexa}]{eitz2012humans}
\bibinfo{author}{Eitz, M.}, \bibinfo{author}{Hays, J.}, \bibinfo{author}{Alexa,
  M.}, \bibinfo{year}{2012}.
\newblock \bibinfo{title}{How do humans sketch objects?}
\newblock \bibinfo{journal}{ACM TOG} .
\bibitem[{Eitz et~al.(2010)Eitz, Hildebrand, Boubekeur and
  Alexa}]{eitz2010sketch}
\bibinfo{author}{Eitz, M.}, \bibinfo{author}{Hildebrand, K.},
  \bibinfo{author}{Boubekeur, T.}, \bibinfo{author}{Alexa, M.},
  \bibinfo{year}{2010}.
\newblock \bibinfo{title}{Sketch-based image retrieval: Benchmark and
  bag-of-features descriptors}.
\newblock \bibinfo{journal}{IEEE TVCG} .
\bibitem[{Federici et~al.(2020)Federici, Dutta, Forr{\'e}, Kushman and
  Akata}]{federici2020learning}
\bibinfo{author}{Federici, M.}, \bibinfo{author}{Dutta, A.},
  \bibinfo{author}{Forr{\'e}, P.}, \bibinfo{author}{Kushman, N.},
  \bibinfo{author}{Akata, Z.}, \bibinfo{year}{2020}.
\newblock \bibinfo{title}{Learning robust representations via multi-view
  information bottleneck}, in: \bibinfo{booktitle}{ICLR}.
\bibitem[{Geirhos et~al.(2018)Geirhos, Rubisch, Michaelis, Bethge, Wichmann and
  Brendel}]{geirhos2018imagenet}
\bibinfo{author}{Geirhos, R.}, \bibinfo{author}{Rubisch, P.},
  \bibinfo{author}{Michaelis, C.}, \bibinfo{author}{Bethge, M.},
  \bibinfo{author}{Wichmann, F.A.}, \bibinfo{author}{Brendel, W.},
  \bibinfo{year}{2018}.
\newblock \bibinfo{title}{Imagenet-trained cnns are biased towards texture;
  increasing shape bias improves accuracy and robustness}.
\newblock \bibinfo{journal}{arXiv preprint arXiv:1811.12231} .
\bibitem[{He et~al.(2016)He, Zhang, Ren and Sun}]{he2016deep}
\bibinfo{author}{He, K.}, \bibinfo{author}{Zhang, X.}, \bibinfo{author}{Ren,
  S.}, \bibinfo{author}{Sun, J.}, \bibinfo{year}{2016}.
\newblock \bibinfo{title}{Deep residual learning for image recognition}, in:
  \bibinfo{booktitle}{CVPR}.
\bibitem[{Hu et~al.(2018)Hu, Shen and Sun}]{hu2018squeeze}
\bibinfo{author}{Hu, J.}, \bibinfo{author}{Shen, L.}, \bibinfo{author}{Sun,
  G.}, \bibinfo{year}{2018}.
\newblock \bibinfo{title}{Squeeze-and-excitation networks}, in:
  \bibinfo{booktitle}{CVPR}.
\bibitem[{Hu and Collomosse(2013)}]{hu2013performance}
\bibinfo{author}{Hu, R.}, \bibinfo{author}{Collomosse, J.},
  \bibinfo{year}{2013}.
\newblock \bibinfo{title}{A performance evaluation of gradient field hog
  descriptor for sketch based image retrieval}.
\newblock \bibinfo{journal}{CVIU} .
\bibitem[{Jiang and Conrath(1997)}]{jiang1997semantic}
\bibinfo{author}{Jiang, J.J.}, \bibinfo{author}{Conrath, D.W.},
  \bibinfo{year}{1997}.
\newblock \bibinfo{title}{Semantic similarity based on corpus statistics and
  lexical taxonomy}.
\newblock \bibinfo{journal}{arXiv} .
\bibitem[{Kampffmeyer et~al.(2019)Kampffmeyer, Chen, Liang, Wang, Zhang and
  Xing}]{kampffmeyer2019rethinking}
\bibinfo{author}{Kampffmeyer, M.}, \bibinfo{author}{Chen, Y.},
  \bibinfo{author}{Liang, X.}, \bibinfo{author}{Wang, H.},
  \bibinfo{author}{Zhang, Y.}, \bibinfo{author}{Xing, E.P.},
  \bibinfo{year}{2019}.
\newblock \bibinfo{title}{Rethinking knowledge graph propagation for zero-shot
  learning}, in: \bibinfo{booktitle}{CVPR}.
\bibitem[{Kipf and Welling(2016)}]{kipf2016semi}
\bibinfo{author}{Kipf, T.N.}, \bibinfo{author}{Welling, M.},
  \bibinfo{year}{2016}.
\newblock \bibinfo{title}{Semi-supervised classification with graph
  convolutional networks}, in: \bibinfo{booktitle}{ICLR}.
\bibitem[{Lei et~al.(2019)Lei, Song, Peng, Ma, Shao and Song}]{lei2019semi}
\bibinfo{author}{Lei, J.}, \bibinfo{author}{Song, Y.}, \bibinfo{author}{Peng,
  B.}, \bibinfo{author}{Ma, Z.}, \bibinfo{author}{Shao, L.},
  \bibinfo{author}{Song, Y.Z.}, \bibinfo{year}{2019}.
\newblock \bibinfo{title}{Semi-heterogeneous three-way joint embedding network
  for sketch-based image retrieval}.
\newblock \bibinfo{journal}{IEEE Transactions on Circuits and Systems for Video
  Technology} \bibinfo{volume}{30}, \bibinfo{pages}{3226--3237}.
\bibitem[{Liu et~al.(2017)Liu, Shen, Shen, Liu and Shao}]{liu2017deep}
\bibinfo{author}{Liu, L.}, \bibinfo{author}{Shen, F.}, \bibinfo{author}{Shen,
  Y.}, \bibinfo{author}{Liu, X.}, \bibinfo{author}{Shao, L.},
  \bibinfo{year}{2017}.
\newblock \bibinfo{title}{Deep sketch hashing: Fast free-hand sketch-based
  image retrieval}, in: \bibinfo{booktitle}{CVPR}.
\bibitem[{Liu et~al.(2019)Liu, Xie, Wang and Yuille}]{liu2019semantic}
\bibinfo{author}{Liu, Q.}, \bibinfo{author}{Xie, L.}, \bibinfo{author}{Wang,
  H.}, \bibinfo{author}{Yuille, A.L.}, \bibinfo{year}{2019}.
\newblock \bibinfo{title}{Semantic-aware knowledge preservation for zero-shot
  sketch-based image retrieval}, in: \bibinfo{booktitle}{ICCV}.
\bibitem[{Lowe(1999)}]{lowe1999object}
\bibinfo{author}{Lowe, D.G.}, \bibinfo{year}{1999}.
\newblock \bibinfo{title}{Object recognition from local scale-invariant
  features}, in: \bibinfo{booktitle}{Proceedings of the seventh IEEE
  international conference on computer vision}.
\bibitem[{Mikolov et~al.(2013)Mikolov, Chen, Corrado and
  Dean}]{mikolov2013efficient}
\bibinfo{author}{Mikolov, T.}, \bibinfo{author}{Chen, K.},
  \bibinfo{author}{Corrado, G.}, \bibinfo{author}{Dean, J.},
  \bibinfo{year}{2013}.
\newblock \bibinfo{title}{Efficient estimation of word representations in
  vector space}.
\newblock \bibinfo{journal}{arXiv} .
\bibitem[{Narasimhan et~al.(2018)Narasimhan, Lazebnik and
  Schwing}]{narasimhan2018out}
\bibinfo{author}{Narasimhan, M.}, \bibinfo{author}{Lazebnik, S.},
  \bibinfo{author}{Schwing, A.}, \bibinfo{year}{2018}.
\newblock \bibinfo{title}{Out of the box: Reasoning with graph convolution nets
  for factual visual question answering}, in: \bibinfo{booktitle}{NeurIPS}.
\bibitem[{Pandey et~al.(2020)Pandey, Mishra, Verma, Mittal and
  Murthy}]{p2020stacked}
\bibinfo{author}{Pandey, A.}, \bibinfo{author}{Mishra, A.},
  \bibinfo{author}{Verma, V.K.}, \bibinfo{author}{Mittal, A.},
  \bibinfo{author}{Murthy, H.}, \bibinfo{year}{2020}.
\newblock \bibinfo{title}{Stacked adversarial network for zero-shot sketch
  based image retrieval}, in: \bibinfo{booktitle}{WACV}.
\bibitem[{Qi et~al.(2016)Qi, Song, Zhang and Liu}]{qi2016sketch}
\bibinfo{author}{Qi, Y.}, \bibinfo{author}{Song, Y.Z.}, \bibinfo{author}{Zhang,
  H.}, \bibinfo{author}{Liu, J.}, \bibinfo{year}{2016}.
\newblock \bibinfo{title}{Sketch-based image retrieval via siamese
  convolutional neural network}, in: \bibinfo{booktitle}{ICIP}.
\bibitem[{Radovanovic et~al.(2010)Radovanovic, Nanopoulos and
  Ivanovic}]{radovanovic2010hubs}
\bibinfo{author}{Radovanovic, M.}, \bibinfo{author}{Nanopoulos, A.},
  \bibinfo{author}{Ivanovic, M.}, \bibinfo{year}{2010}.
\newblock \bibinfo{title}{Hubs in space: Popular nearest neighbors in
  high-dimensional data}.
\newblock \bibinfo{journal}{JMLR} .
\bibitem[{Romera-Paredes and Torr(2015)}]{romera2015embarrassingly}
\bibinfo{author}{Romera-Paredes, B.}, \bibinfo{author}{Torr, P.},
  \bibinfo{year}{2015}.
\newblock \bibinfo{title}{An embarrassingly simple approach to zero-shot
  learning}, in: \bibinfo{booktitle}{ICML}.
\bibitem[{Saavedra(2014)}]{saavedra2014sketch}
\bibinfo{author}{Saavedra, J.M.}, \bibinfo{year}{2014}.
\newblock \bibinfo{title}{Sketch based image retrieval using a soft computation
  of the histogram of edge local orientations (s-helo)}, in:
  \bibinfo{booktitle}{ICIP}.
\bibitem[{Sangkloy et~al.(2016)Sangkloy, Burnell, Ham and
  Hays}]{sangkloy2016sketchy}
\bibinfo{author}{Sangkloy, P.}, \bibinfo{author}{Burnell, N.},
  \bibinfo{author}{Ham, C.}, \bibinfo{author}{Hays, J.}, \bibinfo{year}{2016}.
\newblock \bibinfo{title}{The sketchy database: learning to retrieve badly
  drawn bunnies}.
\newblock \bibinfo{journal}{ACM TOG} .
\bibitem[{Selvaraju et~al.(2017)Selvaraju, Cogswell, Das, Vedantam, Parikh and
  Batra}]{selvaraju2017grad}
\bibinfo{author}{Selvaraju, R.R.}, \bibinfo{author}{Cogswell, M.},
  \bibinfo{author}{Das, A.}, \bibinfo{author}{Vedantam, R.},
  \bibinfo{author}{Parikh, D.}, \bibinfo{author}{Batra, D.},
  \bibinfo{year}{2017}.
\newblock \bibinfo{title}{Grad-cam: Visual explanations from deep networks via
  gradient-based localization}, in: \bibinfo{booktitle}{ICCV}.
\bibitem[{Shen et~al.(2018)Shen, Liu, Shen and Shao}]{shen2018zero}
\bibinfo{author}{Shen, Y.}, \bibinfo{author}{Liu, L.}, \bibinfo{author}{Shen,
  F.}, \bibinfo{author}{Shao, L.}, \bibinfo{year}{2018}.
\newblock \bibinfo{title}{Zero-shot sketch-image hashing}, in:
  \bibinfo{booktitle}{CVPR}.
\bibitem[{Simonyan and Zisserman(2014)}]{simonyan2014very}
\bibinfo{author}{Simonyan, K.}, \bibinfo{author}{Zisserman, A.},
  \bibinfo{year}{2014}.
\newblock \bibinfo{title}{Very deep convolutional networks for large-scale
  image recognition}.
\newblock \bibinfo{journal}{arXiv} .
\bibitem[{Song et~al.(2017)Song, Yu, Song, Xiang and Hospedales}]{song2017deep}
\bibinfo{author}{Song, J.}, \bibinfo{author}{Yu, Q.}, \bibinfo{author}{Song,
  Y.Z.}, \bibinfo{author}{Xiang, T.}, \bibinfo{author}{Hospedales, T.M.},
  \bibinfo{year}{2017}.
\newblock \bibinfo{title}{Deep spatial-semantic attention for fine-grained
  sketch-based image retrieval}, in: \bibinfo{booktitle}{ICCV}.
\bibitem[{Thong et~al.(2020)Thong, Mettes and Snoek}]{thong2020open}
\bibinfo{author}{Thong, W.}, \bibinfo{author}{Mettes, P.},
  \bibinfo{author}{Snoek, C.G.}, \bibinfo{year}{2020}.
\newblock \bibinfo{title}{Open cross-domain visual search}.
\newblock \bibinfo{journal}{CVIU} .
\bibitem[{{Tishby} and {Zaslavsky}(2015)}]{Tishby2015IB}
\bibinfo{author}{{Tishby}, N.}, \bibinfo{author}{{Zaslavsky}, N.},
  \bibinfo{year}{2015}.
\newblock \bibinfo{title}{{Deep Learning and the Information Bottleneck
  Principle}}, in: \bibinfo{booktitle}{ITW}.
\bibitem[{Wang et~al.(2015)Wang, Wang, Yu and Zhang}]{wang2015community}
\bibinfo{author}{Wang, M.}, \bibinfo{author}{Wang, C.}, \bibinfo{author}{Yu,
  J.X.}, \bibinfo{author}{Zhang, J.}, \bibinfo{year}{2015}.
\newblock \bibinfo{title}{Community detection in social networks: an in-depth
  benchmarking study with a procedure-oriented framework}.
\newblock \bibinfo{journal}{VLDBE} .
\bibitem[{Xian et~al.(2018)Xian, Lorenz, Schiele and Akata}]{xian2018feature}
\bibinfo{author}{Xian, Y.}, \bibinfo{author}{Lorenz, T.},
  \bibinfo{author}{Schiele, B.}, \bibinfo{author}{Akata, Z.},
  \bibinfo{year}{2018}.
\newblock \bibinfo{title}{Feature generating networks for zero-shot learning},
  in: \bibinfo{booktitle}{CVPR}.
\bibitem[{Xian et~al.(2017)Xian, Schiele and Akata}]{xian2017zero}
\bibinfo{author}{Xian, Y.}, \bibinfo{author}{Schiele, B.},
  \bibinfo{author}{Akata, Z.}, \bibinfo{year}{2017}.
\newblock \bibinfo{title}{Zero-shot learning-the good, the bad and the ugly},
  in: \bibinfo{booktitle}{CVPR}.
\bibitem[{Xu et~al.(2020a)Xu, Yang, Jiang, Lin, Luo and Xia}]{9070216}
\bibinfo{author}{Xu, F.}, \bibinfo{author}{Yang, W.}, \bibinfo{author}{Jiang,
  T.}, \bibinfo{author}{Lin, S.}, \bibinfo{author}{Luo, H.},
  \bibinfo{author}{Xia, G.S.}, \bibinfo{year}{2020}a.
\newblock \bibinfo{title}{Mental retrieval of remote sensing images via
  adversarial sketch-image feature learning}.
\newblock \bibinfo{journal}{IEEE Transactions on Geoscience and Remote Sensing}
  \bibinfo{volume}{58}, \bibinfo{pages}{7801--7814}.
\newblock \DOIprefix\doi{10.1109/TGRS.2020.2984316}.
\bibitem[{Xu et~al.(2020b)Xu, Yang, Yang and Wang}]{ijcai2020-137}
\bibinfo{author}{Xu, X.}, \bibinfo{author}{Yang, M.}, \bibinfo{author}{Yang,
  Y.}, \bibinfo{author}{Wang, H.}, \bibinfo{year}{2020}b.
\newblock \bibinfo{title}{Progressive domain-independent feature decomposition
  network for zero-shot sketch-based image retrieval}, in:
  \bibinfo{booktitle}{IJCAI}.
\bibitem[{Yang et~al.(2019)Yang, Tang, Zhang and Cai}]{yang2019auto}
\bibinfo{author}{Yang, X.}, \bibinfo{author}{Tang, K.}, \bibinfo{author}{Zhang,
  H.}, \bibinfo{author}{Cai, J.}, \bibinfo{year}{2019}.
\newblock \bibinfo{title}{Auto-encoding scene graphs for image captioning}, in:
  \bibinfo{booktitle}{CVPR}.
\bibitem[{Yelamarthi et~al.(2018)Yelamarthi, Reddy, Mishra and
  Mittal}]{yelamarthi2018zero}
\bibinfo{author}{Yelamarthi, S.K.}, \bibinfo{author}{Reddy, S.K.},
  \bibinfo{author}{Mishra, A.}, \bibinfo{author}{Mittal, A.},
  \bibinfo{year}{2018}.
\newblock \bibinfo{title}{A zero-shot framework for sketch based image
  retrieval}, in: \bibinfo{booktitle}{ECCV}.
\bibitem[{Yu et~al.(2017)Yu, Yang, Liu, Song, Xiang and
  Hospedales}]{yu2017sketch}
\bibinfo{author}{Yu, Q.}, \bibinfo{author}{Yang, Y.}, \bibinfo{author}{Liu,
  F.}, \bibinfo{author}{Song, Y.Z.}, \bibinfo{author}{Xiang, T.},
  \bibinfo{author}{Hospedales, T.M.}, \bibinfo{year}{2017}.
\newblock \bibinfo{title}{Sketch-a-net: A deep neural network that beats
  humans}.
\newblock \bibinfo{journal}{IJCV} .
\bibitem[{Zhang et~al.(2018)Zhang, Shen, Liu, Zhu, Yu, Shao, Shen and
  Van~Gool}]{zhang2018generative}
\bibinfo{author}{Zhang, J.}, \bibinfo{author}{Shen, F.}, \bibinfo{author}{Liu,
  L.}, \bibinfo{author}{Zhu, F.}, \bibinfo{author}{Yu, M.},
  \bibinfo{author}{Shao, L.}, \bibinfo{author}{Shen, H.T.},
  \bibinfo{author}{Van~Gool, L.}, \bibinfo{year}{2018}.
\newblock \bibinfo{title}{Generative domain-migration hashing for
  sketch-to-image retrieval}, in: \bibinfo{booktitle}{ECCV}.
\bibitem[{Zhang et~al.(2020)Zhang, Zhang, Feng, Zhang and Fan}]{zhang2020zero}
\bibinfo{author}{Zhang, Z.}, \bibinfo{author}{Zhang, Y.},
  \bibinfo{author}{Feng, R.}, \bibinfo{author}{Zhang, T.},
  \bibinfo{author}{Fan, W.}, \bibinfo{year}{2020}.
\newblock \bibinfo{title}{Zero-shot sketch-based image retrieval via graph
  convolution network.}, in: \bibinfo{booktitle}{AAAI}.

\end{thebibliography}

\end{document}